\documentclass[conference]{IEEEtran}
\usepackage{cite}
\usepackage{amsmath,amssymb,amsfonts}
\usepackage{graphicx}
\usepackage{textcomp}
\usepackage{url}
\usepackage{pifont} 
\usepackage{soul}
\ifCLASSINFOpdf
\usepackage{amsmath}
\usepackage{wrapfig,blindtext}
\usepackage{algorithmic}
\usepackage{array}
\usepackage{xcolor}
\usepackage[caption=false,font=footnotesize,labelfont=sf,textfont=sf]{subfig}
\usepackage[caption=false,font=footnotesize]{subfig}
\usepackage{fixltx2e}
\usepackage{stfloats}
\usepackage{xspace}

\def\BibTeX{{\rm B\kern-.05em{\sc i\kern-.025em b}\kern-.08em
    T\kern-.1667em\lower.7ex\hbox{E}\kern-.125emX}}
\begin{document}

\title{Optimization Algorithms in Smart Grids: A Systematic Literature Review}

\author{\IEEEauthorblockN{Sidra Aslam, Ala Altaweel, Ali Bou Nassif}
\IEEEauthorblockA{\textit{College of Computing and Informatics} \\
\textit{University of Sharjah}, Sharjah, UAE \\
{[202844, aaltaweel, anassif]}@sharjah.ac.ae}
}

\maketitle
\thispagestyle{plain}
\pagestyle{plain}



\begin{abstract}
Electrical smart grids are units that supply electricity from power plants to the users to yield reduced costs, power failures/loss, and maximized energy management. Smart grids (SGs) are well-known devices due to their exceptional benefits such as bi-directional communication, stability, detection of power failures, and inter-connectivity with appliances for monitoring purposes. SGs are the outcome of different modern applications that are used for managing data and security, i.e., modeling, monitoring, optimization, and/or Artificial Intelligence. Hence, the importance of SGs as a research field is increasing with every passing year. This paper focuses on novel features and applications of smart grids in domestic and industrial sectors. Specifically, we focused on Genetic algorithm, Particle Swarm Optimization, and Grey Wolf Optimization to study the efforts made up till date for maximized energy management and cost minimization in SGs. Therefore, we collected  145 research works (2011 to 2022) in this  systematic literature review. This research work aims to figure out different features and applications of SGs proposed in the last decade and investigate the trends in popularity of SGs for different regions of world. Our finding is that the most popular optimization algorithm being used by researchers to bring forward new solutions for energy management and cost effectiveness in SGs is Particle Swarm Optimization. We also provide a brief overview of objective functions and parameters used in the solutions for energy and cost effectiveness as well as discuss different open research challenges for future research works.
\end{abstract}




\section{Introduction}
\label{sec:introduction}

\begin{figure}[t]
\centering	
\includegraphics[width=0.8\columnwidth]	{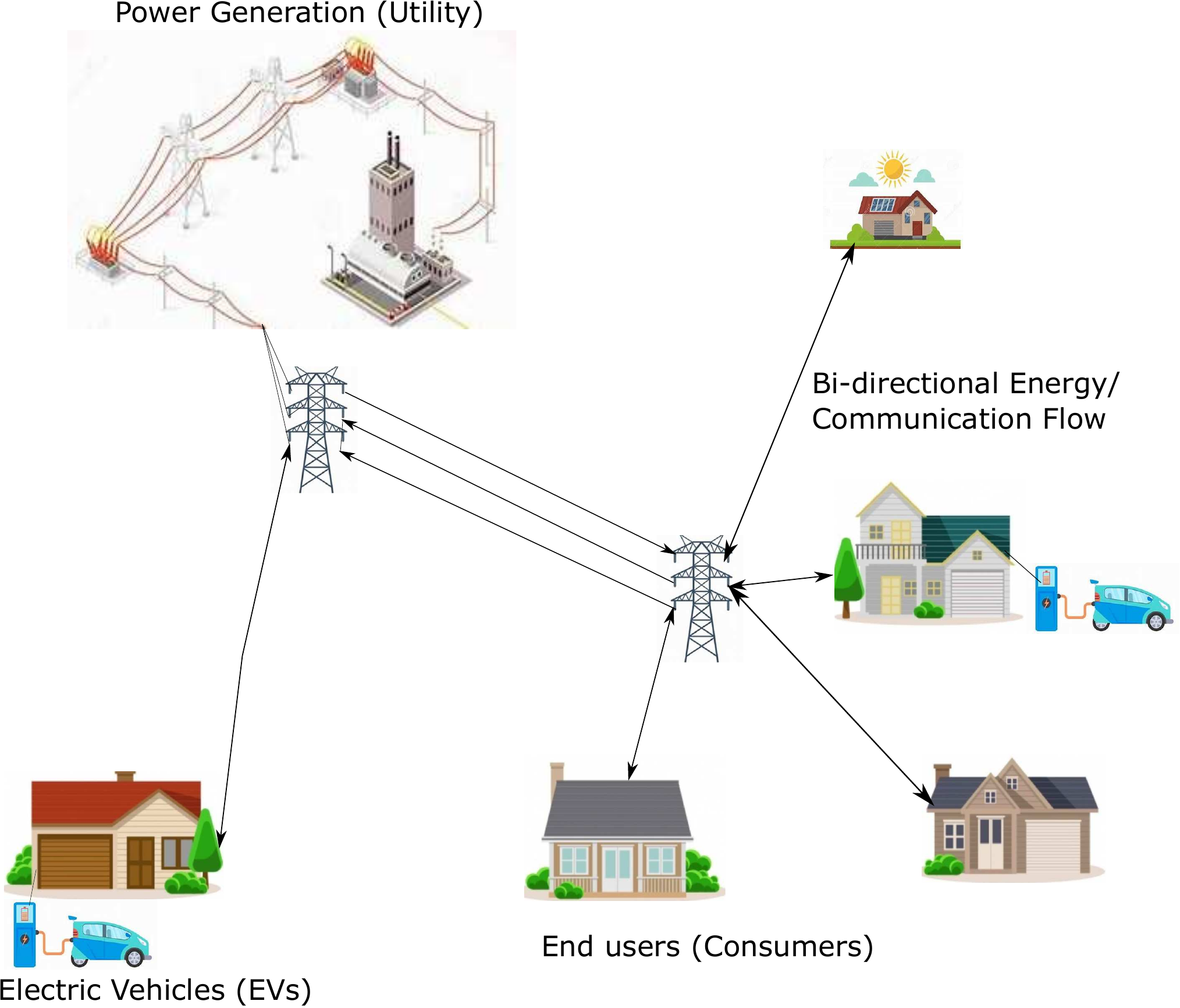}
\caption{A simple smart grid}
\label{smartgrid}
\end{figure}

\begin{figure}[t]
\centering	
\includegraphics[width=0.8\columnwidth]	{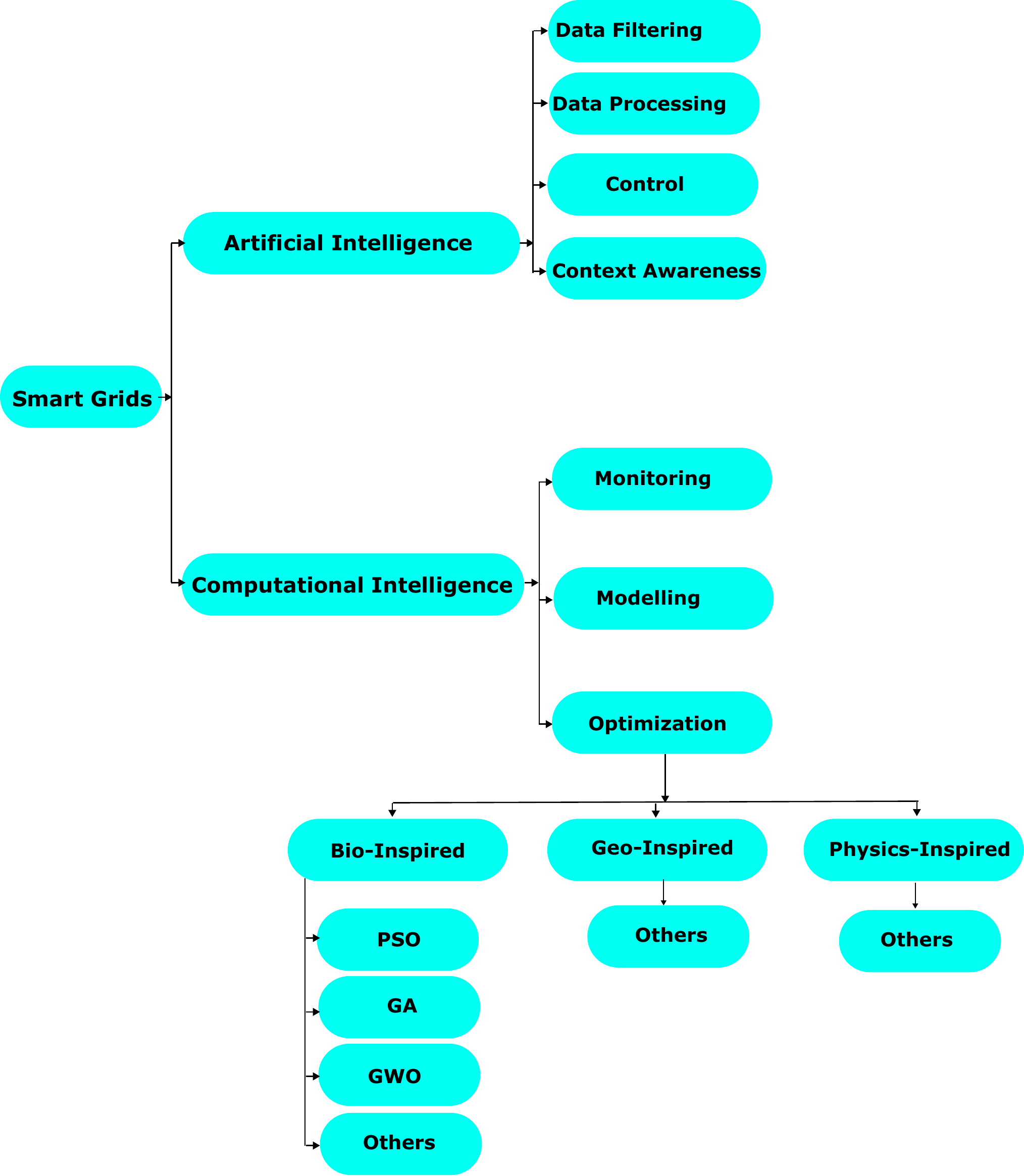}
\caption{Classification of Intelligence techniques in smart grids}
\label{ai}
\end{figure}

Smart grids refers to an electric grid that delivers the electricity from utility (power generator sources/company) to the users (residential/industrial). A simple smart grid connection is shown in Figure~\ref{smartgrid}, with bi-directional communication and power flow between utility (power generation unit) and users (houses/electrical Vehicles). The process of electricity delivery is capable of monitoring, modeling, controlling, data filtering, and data processing with help of number of intelligent features such as Artificial Intelligence (AI) or Computational Intelligence (CI) as shown in Figure~\ref{ai}. SGs allow users to schedule the appliances depending upon pricing hours and its demand that helps in saving energy, increasing reliability, and minimizing costs~\cite{alam2012review,li2010renewable}. Furthermore, SGs support bidirectional power line communications such as Home Area Network (HAN) or Wide Area Network (WAN), and  wireless communications such as ZigBee, 6LowPAN, Z-wave, IoT networks, etc.~\cite{yan2012survey,6011696,wang2017survey,wang2011survey}.

The communications in SGs on one hand have many benefits such as monitoring the load demand, demand response management, while on other hand open the door to threats against secure and profitable electricity and information exchange between the utility and users. The threats that are associated with security (availability, integrity, or confidentiality) are referred to as cyber attacks that result in unauthorized disclosure or monitoring of information on users side (e.g., data required for scheduling appliances)~\cite{8265129,goel2015security}. Many counter attack solutions such as secure data collectors, broadcast authentication, and secure DoS-resistant broadcast authentication protocols have been studied to secure the data collection and coping the demands of users in efficient ways~\cite{uludag2015secure,he2017cyber}. Moreover, besides protecting SGs from certain security attacks, other challenges are faced by both utility and users (energy supply and energy demand) such as energy management, cost efficiency, reducing power losses, and reducing pollutant emissions~\cite{moretti2017systematic,clastres2011smart}. The aforementioned challenges can be addressed using optimization techniques in SGs to maximize the profit (for both users and utility) by managing electricity distribution and reducing emissions.

Optimization in SGs is employed to find the conditions with maximum benefits while (at the same time) minimizing the electricity wastage and cost~\cite{dong2012distributed}. Hence, optimization problem in SGs is defined as a scenario (i.e., an objective function) that has to be minimized or maximized while (in most cases) influenced by a set of variables and/or constraints. Based on the objective function and given constraints, optimization problems can be classified into integer programming, mixed integer non-linear, linear, and non-linear programming problems~\cite{rao2019engineering,6861959}. To find the optimal solution for the objective function, a variety of techniques have been proposed. One of the major techniques are ``Metaheuristic'' optimization techniques, that are considered as a major solver to many optimization problems. Some variations of the Metaheuristic optimization techniques are inspired from bio, geo, or physics fields and are further divided into different search algorithms for obtaining the optimal solution. As shown in Figure~\ref{ai}, the classification of optimization techniques in SG are grouped under computational intelligence. There are many examples of the aforementioned algorithms such as Particle Swarm Optimization (PSO), Grey Wolf Optimization (GWO), Differential-Evolution (DE), Genetic Algorithm (GA), Artificial-Bee Colony (ABC), Ant-Colony Optimization (ACO), etc.~\cite{6861959,minai2021metaheuristics}. As optimization of SGs is a leading topic of research that helps in reducing the major problems of the energy sector by providing promising solutions~\cite{reddy2017review,hota2014issues,marah2018algorithms,kolokotsa2019integration}, we conduct a systematic review in this paper with a focus on PSO, GA, and GWO as the scope of our survey. In Table~\ref{Abbreviations_Keywords}, we present the abbreviations and keywords for the used terms in our paper.

The main objective of this systematic literature review is to enhance the research community's understanding about optimization algorithms that are employed to propose novel applications focusing on cost efficiency and energy management of SGs. We focused on PSO, GA, and GWO to compare the basic, easy to implement, and flexible bio-inspired meta-heuristic techniques and study their benefits for the environment, utility, and users. Our comparison aims to demonstrate the evolution of optimization techniques, their rate of convergence, delivering most optimized results, and to find out which algorithm is flexible and provide (most/near) to optimized results. For future work, we aim to expand our research for other optimization algorithms (i.e., ABC, ACO). Our contributions in this paper are:

\begin{itemize}
 
\item A brief review of concepts and applications of optimization algorithms (PSO, GA, and GWO) and terms used in smart grids (Section~\ref{Background}).

\item We review novel applications and features of smart grids with supporting research work published between 2011-2022 (Section~\ref{Results&Discussion}).

\item We collect certain statistics from different sources to show smart grids as an emerging field of interest and investment by different countries around the world (Section~\ref{Results&Discussion}).

\item We study the research works for cost minimization and energy management in smart grids using GA, PSO, and GWO. The related work is gathered in form of tables, highlighting the objective functions, parameters, and constraints that influence the objective functions (Section~\ref{Results&Discussion}).

\item We also highlight the commonly used optimization algorithm among GA, PSO, and GWO that is being used for deriving solutions in efficient cost and energy management in smart grids (Section~\ref{Results&Discussion}).

\item We discuss the open research challenges for optimizing the cost-effective energy management, monitoring the cyber and physical attacks with most advanced optimization algorithms (Section~\ref{opc}).

\end{itemize}

\begin{table}[t]
\centering
\caption{Abbreviations and Keywords for: Optimization Algorithms, Parameters, and Terms.}
\label{Abbreviations_Keywords}
\begin{tabular}{|l|l|}
\hline
\multicolumn{2}{|c|}{\textbf{Optimization Algorithms}} \\
\hline
GA        & Genetic Algorithm                  \\ \hline
PSO       & Particle Swarm Optimization                           \\ \hline
GWO       & Grey Wolf Optimization                 \\ \hline

\multicolumn{2}{|c|}{\textbf{Terms and Parameters}} \\
\hline
SGs         & Smart Grid(s)                         \\ \hline    
MG          & Micro Grid(s)                           \\ \hline   
DG          & Distributed Generation                     \\ \hline
DR          & Demand Response               \\ \hline   
PAR         & Peak to Average Ratio                          \\ \hline 
SLR         & Systematic Literature Review                        \\ \hline 
AI          & Artificial Intelligence\\ \hline
SVM         & Support Vector Machine \\ \hline
SM          & Smart Meters \\ \hline
DSM         & Demand Side Management \\ \hline
HEM         & Home Energy Management \\ \hline
EV          & Electric Vehicles \\ \hline


\end{tabular}
\end{table}

\section{Background}
\label{Background}

Digital power systems were first introduced by~\cite{luqiang1999basic} in last decade of 20 century. With time, more advancements were accomplished to make the systems more reliable and sustainable for interconnected power grid systems at larger area (or smart grid). The definition of SGs can be varied according to their usage, focused characteristics (i.e., cost, renewable resources, management, security, or used applications to make SGs), and source of energy (electricity or gas). In our paper, we focus on electrical SGs, their costs and energy management. Hence, we define a smart grid as: 

``\textit{A set of intelligent and inter-connected (bi-directional, i.e., IoT) networks and applications (i.e., AI, Machine Learning (ML)) working together to provide feasible usage of electricity (according to consumer demand), maintain sustainability, and ensure/maximize the economical power supply from utility to the user end.} '' 

Depending upon the area of deployed SGs (population, region, or weather), authorities (private sectors, governments, etc.,) that control/budget power resources, their information related, and user/utility demand/reservoirs, the characteristics of SGs might vary and are improved accordingly. For instance, USA uses the demand response from users to manage the electricity wastage, intelligent systems having self-healing powers (that adjusts itself according to power outage/disturbance) that provides security against cyber/physical attacks~\cite{en5051321}. Moreover, owing to the importance of SGs, many developed and under-developed countries are investing in the field and utilizing the knowledge of distributed computing on real-time simulation to enhance the research effort of SGs and provide beneficial solutions to the regions specific problems. South Korea~\cite{mah2012governing,kim2010appropriate}, China~\cite{yuan2014smart,zhu2015past}, New Zealand~\cite{nair2009smartgrid,stephenson2018smart}, North America~\cite{wang2011case}, Pakistan~\cite{irfan2017opportunities}, UAE \& other Gulf countries~\cite{bayram2016overview}, and United Kingdom~\cite{connor2014policy} are some examples of countries, leading in research field on SGs. 

With more growing benefits of using SGs (such as coping up with increased demand of electricity), researchers have pointed out many challenges that users/utility or environment have/had faced. The challenges are related to environment well-being (carbon dioxide emissions), security (cyber and physical attacks), energy management, and cost effectiveness. Nevertheless, for all aforementioned  problems/challenges, optimization of SGs can provide significant solutions.

\subsection{Technical Background} 

In this section, we provide a precise overview of certain technical terminologies used in our survey article.
 
\textit{Micro Grids (MG)} provides freedom of usage to users when connected/non-connected to grids~\cite{jian2009application}.~\textit{Demand Response (DR)} is known as the changing demand of electricity usage by the end-users (depending upon the price of electricity during peak/lean hours, routine/daily patterns of electricity usage). DR helps in monitoring the usage of electricity and provides feasibility to the end-users for energy management in peak hours~\cite{lee2013assessment,borenstein2002dynamic}. Distributed Generation (DG) is used for the advanced technologies that help in generating electricity (at or near end-user/consumer facility). DG includes solar panels, wind turbines, hydro-power plants, etc.~\cite{DistributedGeneration}.~\textit{Optimization} is a technique used to get the best fit solution for a given problem, either by maximizing or minimizing a given objective function. There are several optimization algorithms used for optimization in SGs~\cite{ahat2013smart}.  In this systematic literature review article, we focus on Genetic Algorithm (GA)~\cite{holland1992adaptation}, Particle Swarm Optimization (PSO)~\cite{poli2007particle}, and Grey Wolf Optimization~\cite{mirjalili2014grey} algorithm.

\begin{figure}[t]
\centering	
\includegraphics[width=0.4\columnwidth]	{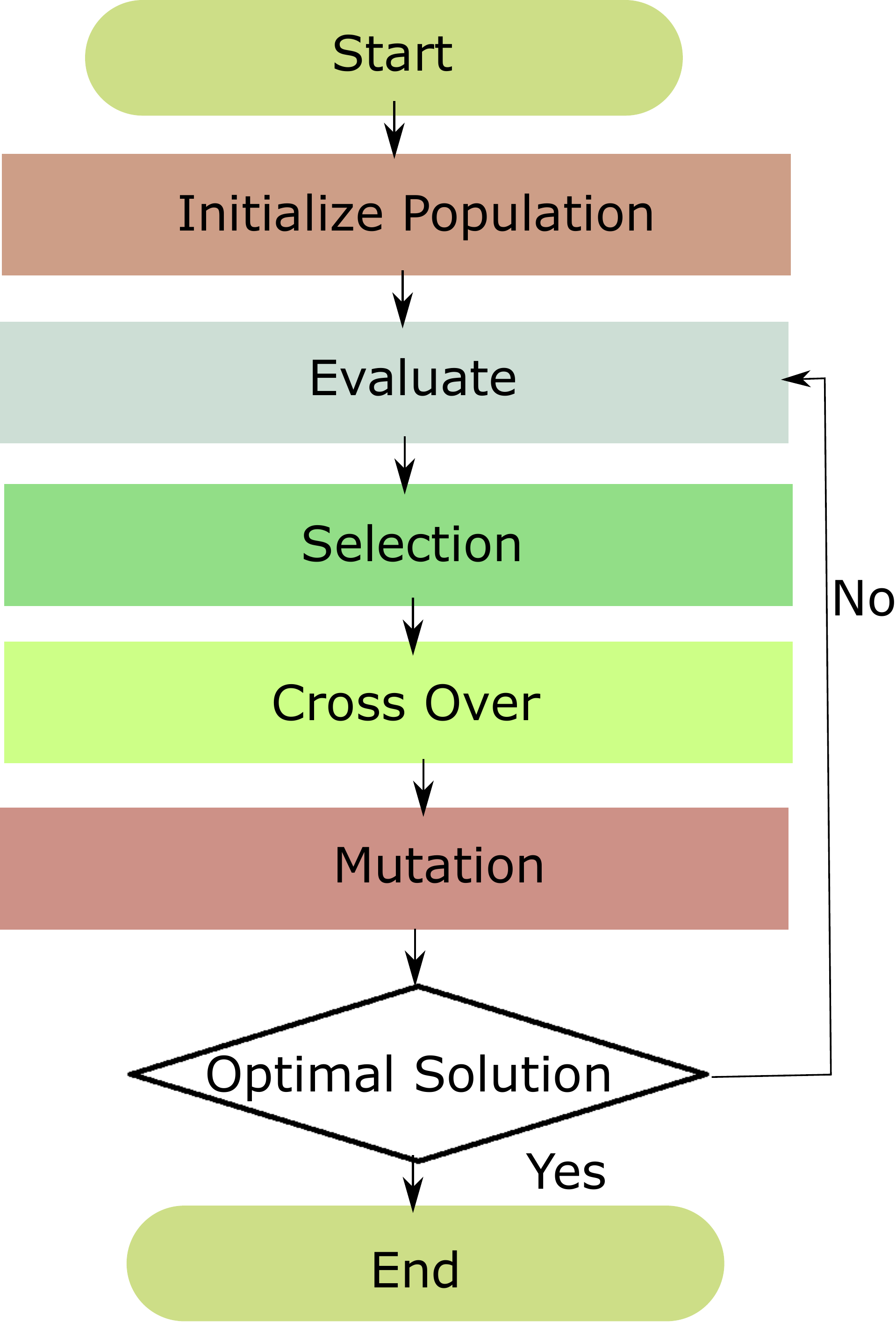}
\caption{Flow chart of genetic algorithm }
\label{geneticalgo}
\end{figure}

\textbf{Genetic Algorithm (GA)} is a bio-inspired optimization algorithm based on the concept of biological genetics~\cite{holland1992adaptation}. Basically, the natural phenomena of division of genes (from parents), the probability of receiving the genes (in off-springs) are combined in this search algorithm to find the best/optimal solution~\cite{kumar2010genetic}. The algorithm uses a group of individuals (population) and their characteristics (chromosomes) to find the fittest individual. The process involves crossing over and mutation of genes (single trait/value from a chromosome) for mixing the traits between the parents. Each population of parents reproduces and form another population of off-springs with more fitness values and mutations than their parents. This evolutionary algorithm improves the parameter values in every iteration, and the process is repeated until desired end-conditions (optimal values) are satisfied/met. Figure~\ref{geneticalgo} shows the flow-chart of Genetic algorithm. The algorithm is initialized by a population of chromosomes or a set of possible solutions called generation. The generation consists of chromosomes from every individual and are evaluated to gain its quality. From a random selection of individuals from population, the characteristics (chromosomes) are altered  using cross over and mutation for finding optimal solutions. The process is repeated until the desired results are gained without any weak characteristic in chromosome.

\begin{figure}[t]
\centering	
\includegraphics[width=0.5\columnwidth]	{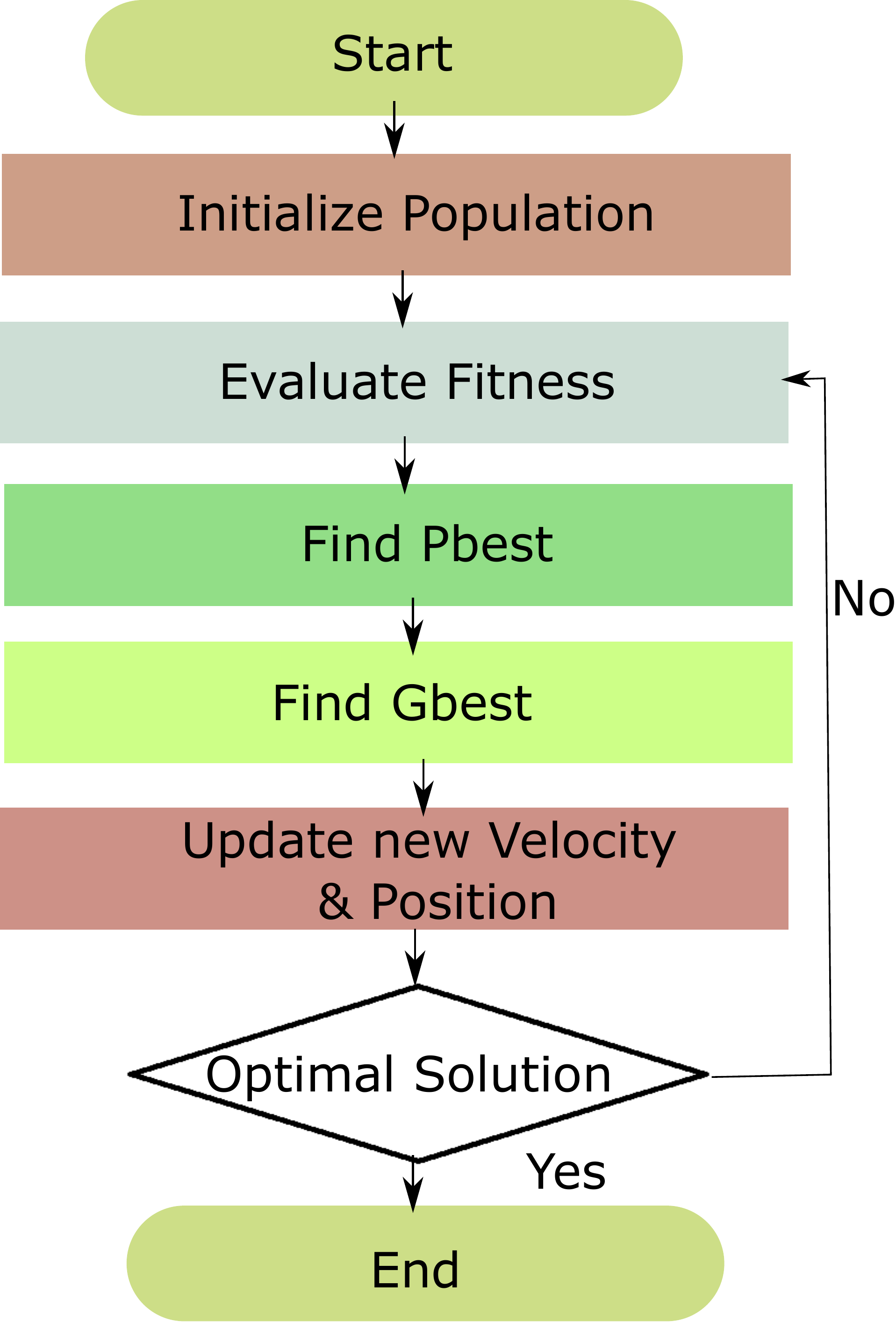}
\caption{Flow chart of particle swarm optimization algorithm }
\label{psoalgo}
\end{figure}

\textbf{Particle Swarm Optimization (PSO)} is inspired from swarm's behavior and intelligence to make optimization decisions~\cite{kennedy1995particle,poli2007particle}. In PSO, every individual swarm from a population of swarms moves with a certain velocity to find the optimal solution. Figure~\ref{psoalgo} illustrates the flow-chart of PSO (i.e., starting from the initialization of swarm population up to finding the optimal solution). The swarm population consists of individual particles/swarms moving with some velocity to search for the best suitable solution according to given objective function. The algorithm evaluates the fitness of each particle and finds the best particle as best solution (Pbest) and global optimal solution (Gbest) is updated after every iteration as optimal solution from the population. Multi-Objective PSO (MoPSO) is used to solve multi-objective problems, while PSO is used for one objective function only. 

\begin{figure}[t]
\centering	
\includegraphics[width=0.5\columnwidth]	{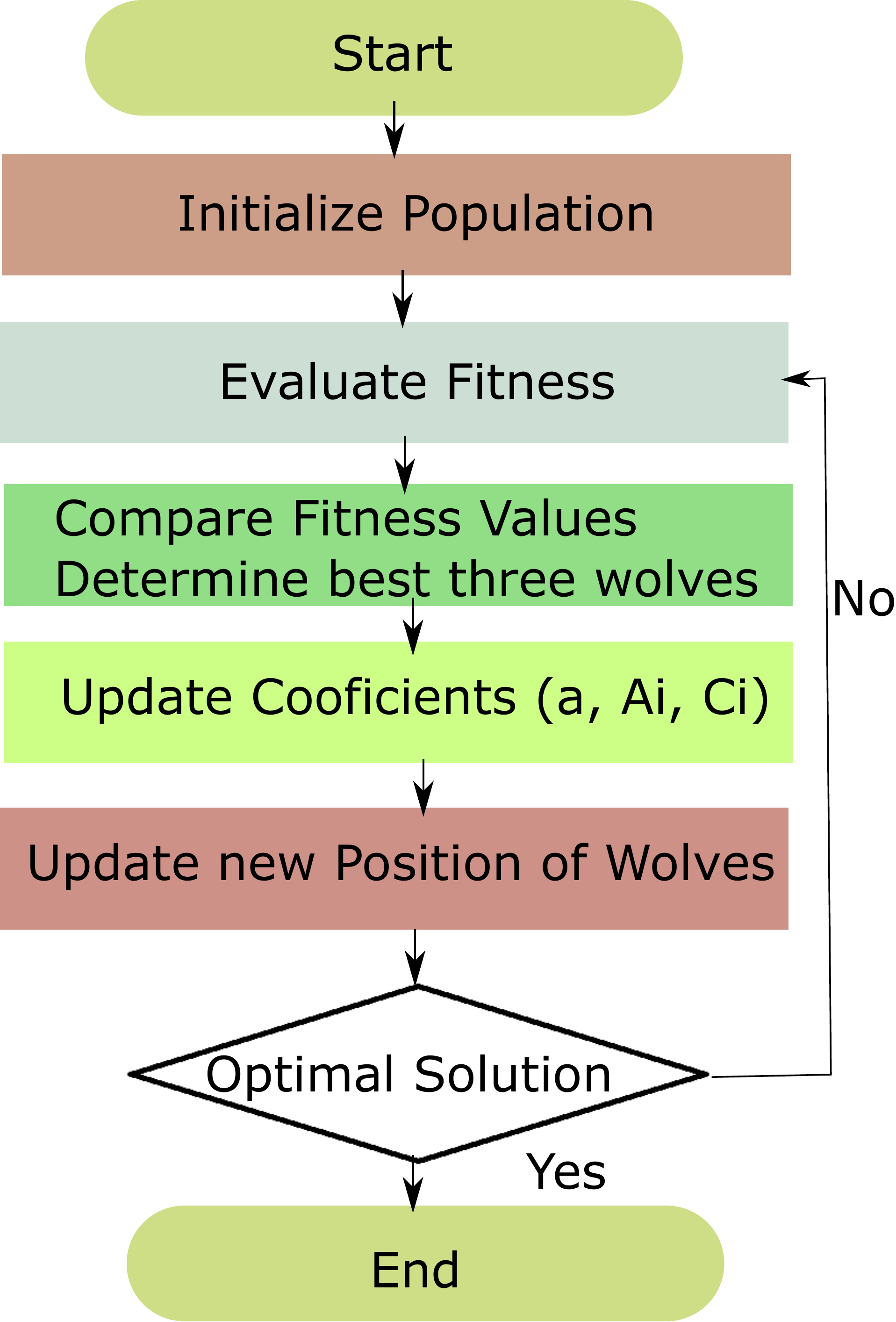}
\caption{Flow chart of grey wolf optimization}
\label{gwoalgo}
\end{figure}

\textbf{Grey Wolf Optimization (GWO)} is another bio-inspired technique to find optimal solution for a given objective function~\cite{mirjalili2014grey}. GWO uses the hunting nature and social leadership characteristics of grey wolves. Four hierarchical groups (Alpha, Beta, Delta, and Omega) of wolves are used to track and attack the prey. The process of finding the optimal solution is defined according to wolf's group characteristics of moving, chasing, and/or attacking the prey. Figure~\ref{gwoalgo} shows the flow-chart for working of GWO, starting with the initialization of the wolves population for every group of grey wolves e.g., Alpha, Beta, Delta, and Omega. Beta, Delta, and Omega chase for the optimal solution and determines the best three values (wolves) for a given optimization problem. The tuning vector coefficients, A and C, determine the best wolf in the herd. \textit{C} can be changed to any random value in every iteration to explore more solutions to find the best position of wolf. While, \textit{A} is dependent upon values of \textit{a}, such that -1 $\leq$ \textit{a} $\leq$ 1 to attack the prey. The changing values of \textit{A} helps in moving of wolves in opposite directions (east/west, as \textit{a} changes from -1 to 1) from prey centered at 0. Whereas, \textit{i} indicates the specific wolf number from a group. With the help of this approach, wolves can search globally for optimal solutions. Once an optimal solution is found, wolves update their coefficients and positions accordingly.


\section{Literature Review}

Energy management is important for every aspect of life for minimizing the energy losses and providing ecological benefits to users and utilities. The Demand Side Management (DSM), utility energy control, load shifting, and cost analysis for utility/user are important factors that control and influence the energy management~\cite{article}. DSM allows users to reduce the power/energy loss (due to Joules effect). Load shifting enhances the privacy of the consumed loads by flattening the energy peaks at user side. Using the comparison of total energy loss or the carbon emissions with the cost paid for the energy usage helps to urge the importance of energy management in SGs. An average household in Italy under specific scenarios can save from 35,847 KWh to 5,516,361 KWh~\cite{article}. Furthermore, SG possess several benefits like diversiﬁcation of the energy resources supply, time deferral of investments, electrical energy loss reduction, etc.

A comparative analysis of various optimization techniques for energy management in SG was provided in~\cite{malik2019towards}. The authors suggested that a secure and efficient grid is required for efficient communication, data exchange among different communication modules, and optimization. By using proper optimization skills and well-distributed/maintained power/energy resources, the utilization losses can be reduced. Moreover, reliability of grid can be achieved as well as pollutant emissions and energy costs can be minimized. The authors reviewed GA, ACO, PSO, Neural Network, and Memetic algorithms for optimization of energy and costs in SG. Their conclusion is that employing optimization algorithms in SG is still at its early stage and therefore more works/efforts have to be done in this research field.

In order to reduce uncertainty and enhance the stability in SGs, probabilistic optimization is the most popular and advanced technique used for Home Energy Management (HEM), DSM, microgrid integration, and economic dispatch problems~\cite{riaz2022probabilistic}. The authors of~\cite{riaz2022probabilistic} reviewed a study of different probabilistic optimization techniques in smart power system and classified them into several categories. Specifically, stochastic optimization, robust optimization, distributionally robust optimization, and chance constraints optimization. The analysis of the aforementioned classes provided a reliable solution with high computational cost, robust optimization, and offered more conservative solutions but has lower computational cost. Whereas, the distributionally robust optimization as compared to the robust technique is less conservative. Nevertheless, robust optimization is the most popular optimization technique for energy management in SGs. HEM system in SGs helps in monitoring and optimizing the energy flows from utility to users considering the power storage and smart appliances. The two most prominent methods for optimal HEM system are Model Predictive Control (MPC) and Reinforcement Learning (RL). The authors of~\cite{Kallio_2022} conducted a comparative review of MPC and RL methods in dynamic environment. Moreover, a case study was also performed using four DualSun Photo Voltic panels. The studies were conducted for two days and concluded that modeling and forecasting are the two major disadvantages of MPC. However, RL being model-free method is more feasible to use as compared to MPC, however, in specific cases, it can lead to unstable control. 

A review for evaluating the performance of three common Artificial Intelligence load forecasting methods (i.e., long short-term memory, group method of data handling, and adaptive neurofuzzy inference system) was presented in~\cite{salehimehr2022short}. Long short-term memory model was also evaluated under noisy condition. The data sets were divided into training and testing sets that were used for studying the impacts of the three methods based on mean and Root Mean Square Error (RMSE) calculated using MATLAB. The results concluded that long short-term memory model has more ability for data handling as compared to the other two methods. A systematic literature review was conducted for smart energy conservation system (2013-2019)~\cite{KIM2021110755}. The review suggests that constructing new infrastructure, adopting newly developed strategies, and energy monitoring at SGs could help in energy management in smart cities. Moreover, application of energy management methods and implementing newly developed energy management technologies are promising solutions for energy conservation.

To the best of our knowledge, cost and energy management in SGs optimized by GA, PSO, and GWO algorithms altogether have never been the scope (focus) of systematic literature review using~\cite{Kitchenham07guidelinesfor} guidelines. This literature review is novel as it provides the quality papers from four research databases (published between 2011-2022) that bring forward novel ideas for applications and features for optimizing results and also elaborates the importance of the SGs. We believe that the outcomes of our survey are crucial for the research community as it follows unbiased and well-known guidelines in conducting SLR.  


\section{Literature Review: Methodology}
\label{Survey_Methodology}

\begin{figure}[t]
\centering	
\includegraphics[width=0.7\columnwidth]	{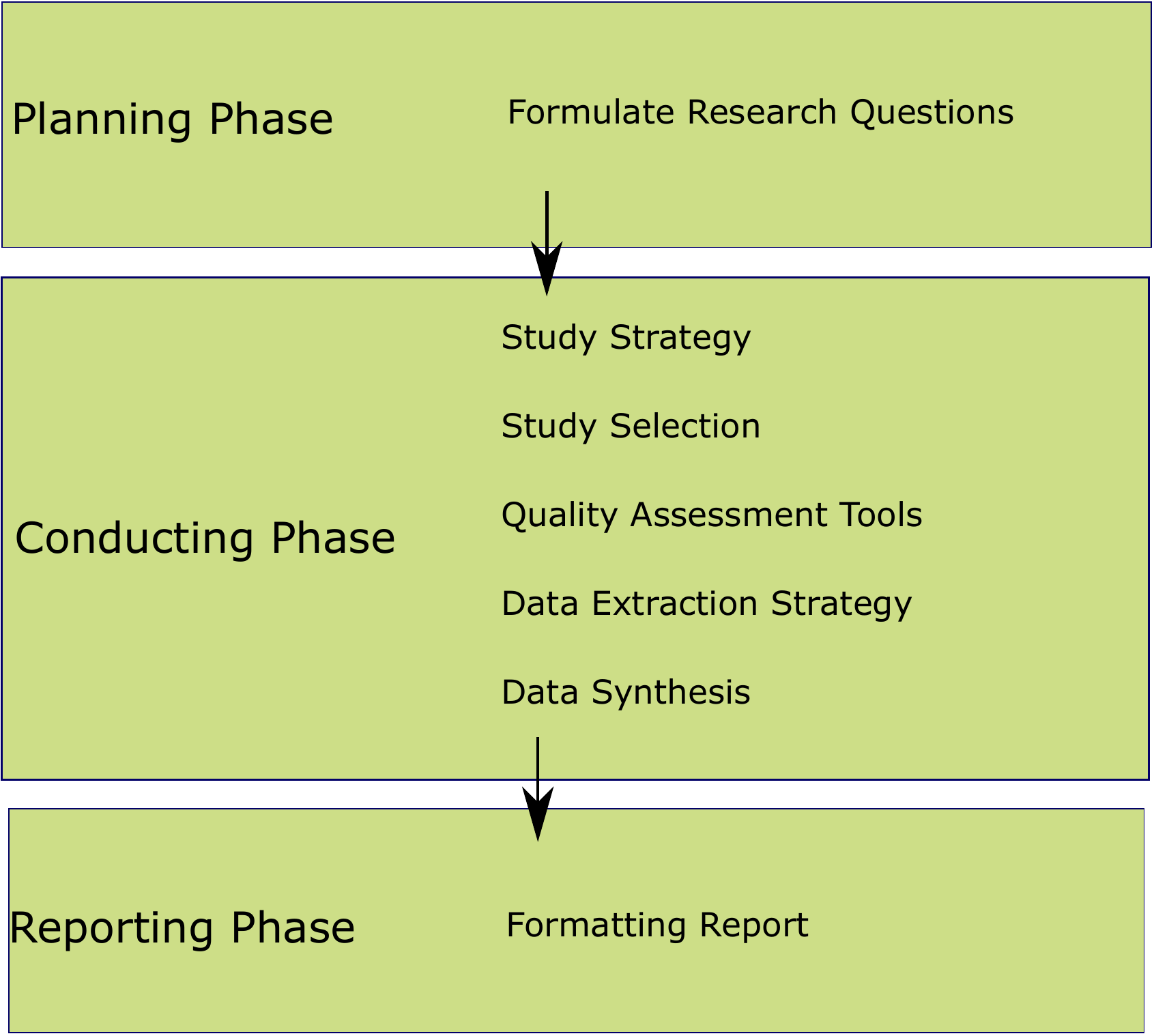}
\caption{Kitchenham and Charters systematic literature review phases~\cite{Kitchenham07guidelinesfor}}
\label{slr}
\end{figure}

Kitchenham and Charters in 2007 presented guidelines for performing Systematic Literature Reviews (SLR) in Software Engineering~\cite{Kitchenham07guidelinesfor}. The guidelines can be implemented in variety of research areas for conducting SLR. Figure~\ref{slr} shows the three phases that are divided into sub-phases of the review. The first phase is planning in which some research questions are formulated to bring forward the motivation/cause of the SLR. The second phase is conducting phase in which search process is performed for getting answers of the formulated research questions using specified terms, criteria, and assessment tools. Finally, the reporting phase is used to compile the gathered data (i.e., research articles) into various forms (e.g., text, graphs, tables, and/or figures). In the following sections, we provide a detailed search methodology for our survey article.

\subsection{Planning Phase}
\label{Planning_Phase}

The Planning phase is the first step in SLR, in which we formulate the research questions to motivate our future research area. Following are the research questions drafted for our survey:

\begin{itemize} 
    \item {\textbf{RQ1:} What are the applications of Smart Grids?}
    \item {\textbf{RQ2:} Why Smart Grid is a crucial and promising research area?}
    \item {\textbf{RQ3:} What are the objective functions/parameters used for obtaining an optimal energy management/cost-efficient system in Smart Grids?}
    \item {\textbf{RQ4:} How much GA, GWO, and PSO algorithms have been employed to enhance energy and cost handling in Smart Grids?}
\end{itemize}

\subsection{Conducting Phase}
\label{Conducting_Phase}

During the conducting phase, various steps (defined below) were followed to gather the useful data and answer the formulated research questions in Section~\ref{Planning_Phase}.  

\subsubsection{Search Strategy}
\label{Search_Strategy}

The data collection is based on the term-based selective search within the articles from different resources from 2011 to 2022. We used Google scholar, IEEE, Elsevier, and Springer research databases to search the defined terms. The research questions were used for identification of research terms, searched from ``tittle'' and ``abstract'' of the data (i.e., research articles). Also, benefiting from advanced search tools like ``AND'' and ``OR'' to filter the research results.

\subsubsection{Search Resources and Search Terms}
\label{Search_Resources}

We consider following research resources for reporting the research articles (i.e., journals and conference papers):

\begin{enumerate}
     \item {Google Scholar}
     \item {IEEE Explorer}
     \item {Elsevier/Science Direct}
     \item {Springer}
\end{enumerate}

Following key terms are applied on IEEE Explorer to gain results from 2011 to 2022:

\begin{itemize}
    \item  Genetic algorithm \textit{AND} smart grid \textit{AND} AI.
    \item  Particle swarm optimization \textit{AND} smart grid \textit{AND} AI.
    \item  Grey Wolf Optimization \textit{AND} smart grid \textit{AND} AI.
\end{itemize}

Following key terms are applied on Google Scholar and Springer to gain results from 2011 to 2022 (\textit{With all of the words}, \textit{Exact Phrase}, and \textit{At least one word} are search options):

\begin{itemize}
    \item  \textit{With all of the words:} Smart grids \textit{Exact Phrase}: Genetic algorithm, \textit{At least one word}: AI.
    \item  \textit{With all of the words:} Smart grids \textit{Exact Phrase}: Particle swarm algorithm, \textit{At least one word}: AI.
    \item  \textit{With all of the words:} Smart grids \textit{Exact Phrase}: Grey wolf algorithm \textit{At least one word}: AI.

\end{itemize}
Following key terms are applied on ScienceDirect/Elsevier to gain results from 2011 to 2022:
\begin{itemize}
    \item  Genetic algorithm in smart grids using AI (ScienceDirect/Elsevier).
    \item  Particle swarm optimization in smart grids using AI (ScienceDirect/Elsevier).
    \item  Grey Wolf Optimization in smart grids using AI (ScienceDirect/Elsevier).
\end{itemize}

Following are key words for applications of smart grids used in the four research resources:
\begin{itemize}
    \item Novel Applications of Smart Grids.
\end{itemize}

We got a total of~$\sim$107 relevant articles from IEEE,~$\sim$21 from Springer, and more than~$\sim$50 each from Google Scholar and Elsevier. Since the gained research works were very large in numbers and some of them were off-topic, we removed the off-topic, redundant, and less informative research works using the study selection phase of SLR that we will present in Section~\ref{studyselection} below.

\subsubsection{Study Selection}
\label{studyselection}

This section states the techniques of study selection applied to pull all raw and off-topic articles from the large research results. Hence, inclusion/exclusion criteria were developed as filtering mechanisms to find accurate and precise answers for our research questions (presented in Section~\ref{Planning_Phase}). The exclusion and inclusion criteria are:

\begin{itemize}
    \item \textbf{Inclusion:} Articles related cost-effectiveness and energy management through GA/PSO/GWO techniques in Smart grids (learned from skimming through the paper).
    \item \textbf{Inclusion:} Articles published in English language only.
    \item \textbf{Inclusion:} Scopus and Scimago indexed journals and conference papers.
    \item \textbf{Inclusion:} Learning the optimization in basic (not enhanced) versions of GA, PSO, and GWO.
    
    \item \textbf{Exclusion:} Books and book-chapters.
    \item \textbf{Exclusion:} Articles repeating the idea or giving solution to the same problems/issues but with different techniques other than optimization of smart grids.
    \item \textbf{Exclusion:} All types of review papers.
    \item \textbf{Exclusion:} Research papers that are repetitive, define the impacts, or case studies of already published articles.
\end{itemize}

After applying the inclusion and exclusion criteria, the gathered data was reduced to $\sim$150 research papers and the resultant information was refined on basis of quality assessment tools that we present in Section~\ref{Quality_Assessment_Tools} below.

\subsubsection{Quality Assessment Tools}
\label{Quality_Assessment_Tools}

To evaluate the quality of gathered data from all research resources, quality assessment tools were used to refine the search process. The quality assessment tools were applied to cover only the most relevant papers from the selected $\sim$150 papers. In this step, a total of 6 quality assessment questions were formulated. These quality assessment questions are essential for scoring and assessing the quality of the research papers. Each paper was scored against all questions on the scale of 0 to 1. If the paper fully answers the quality assessment questions then it was given the score 1. If the research paper partially answers the quality assessment question, it was marked 0.5 and 0 was given for incomplete or vague ideas to answer the quality assessment question. From 6 quality assessment questions, if the paper scored 3 out of 6, we included it in our SLR, otherwise, it is excluded. We applied strict ratio of quality assessment tools to include quality articles from a huge ratio of published papers in the field. In the following list, we present the quality assessment questions that we applied:

 \begin{itemize}
   \item \textbf{Q1:} Does the paper clearly defines the motivation and contributions?
   \item \textbf{Q2:} Does the paper clearly defines the objective function(s) and parameters?
   \item \textbf{Q3:} Are there any clear and useful reported results after applying the optimization/proposed algorithm?
   \item \textbf{Q4:} Are the results provided in the paper, compared to benchmarks (previously proposed techniques for same problem statement)?
   \item \textbf{Q5:} Is the problem statement helpful in monitoring the energy management and reducing cost functions or applications of Smart grid?
   \item \textbf{Q6:} Is the conducted study in the research article practically helpful and can be implemented (i.e., in industry or real-world scenarios)?
 \end{itemize}

\subsubsection{Data Extraction Strategy and Data Synthesis}

In this step, the final collected list of papers was used to extract the information required to answer the research questions in Section~\ref{Planning_Phase}. The following set of information is extracted:

\begin{itemize}
    \item  Article's tittle and its publication year.
    \item  Proposed objective function and optimization parameters.
    \item  The assumptions made in proposed solutions.
    \item  Publisher of the article.
    \item  Proposed solutions for energy management and cost effectiveness in Smart grids.
    \item  State-of-art proposed applications of Smart grids.
  
\end{itemize}
 
The gathered data was synthesised in form of graphs, tables, figures, and text in enhanced and informative ways to address the research questions.

\subsection{Reporting Review}
After gathering the data successfully from the aforementioned sections, the review is reported carefully and precisely answering each research question.

\section{Results and Discussion}
\label{Results&Discussion}

This section answers the research questions listed in~\ref{Planning_Phase}.

\subsection{RQ1: What are the applications of Smart Grids?}

\begin{figure}[t]
\centering	
\includegraphics[width=0.8\columnwidth]	{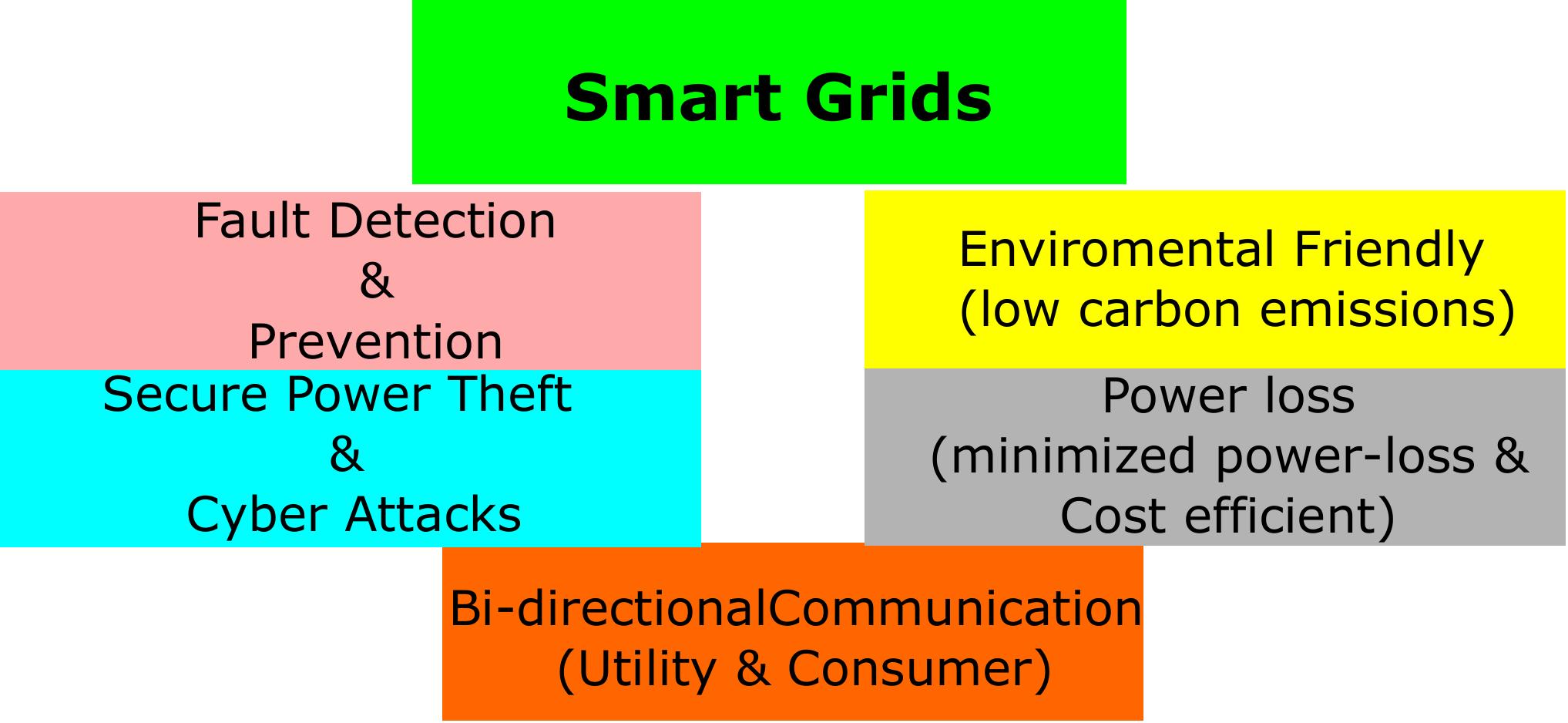}
\caption{Features of smart grids }
\label{advsg}
\end{figure}

\begin{figure}[t]
\centering	
\includegraphics[width=0.9\columnwidth]	{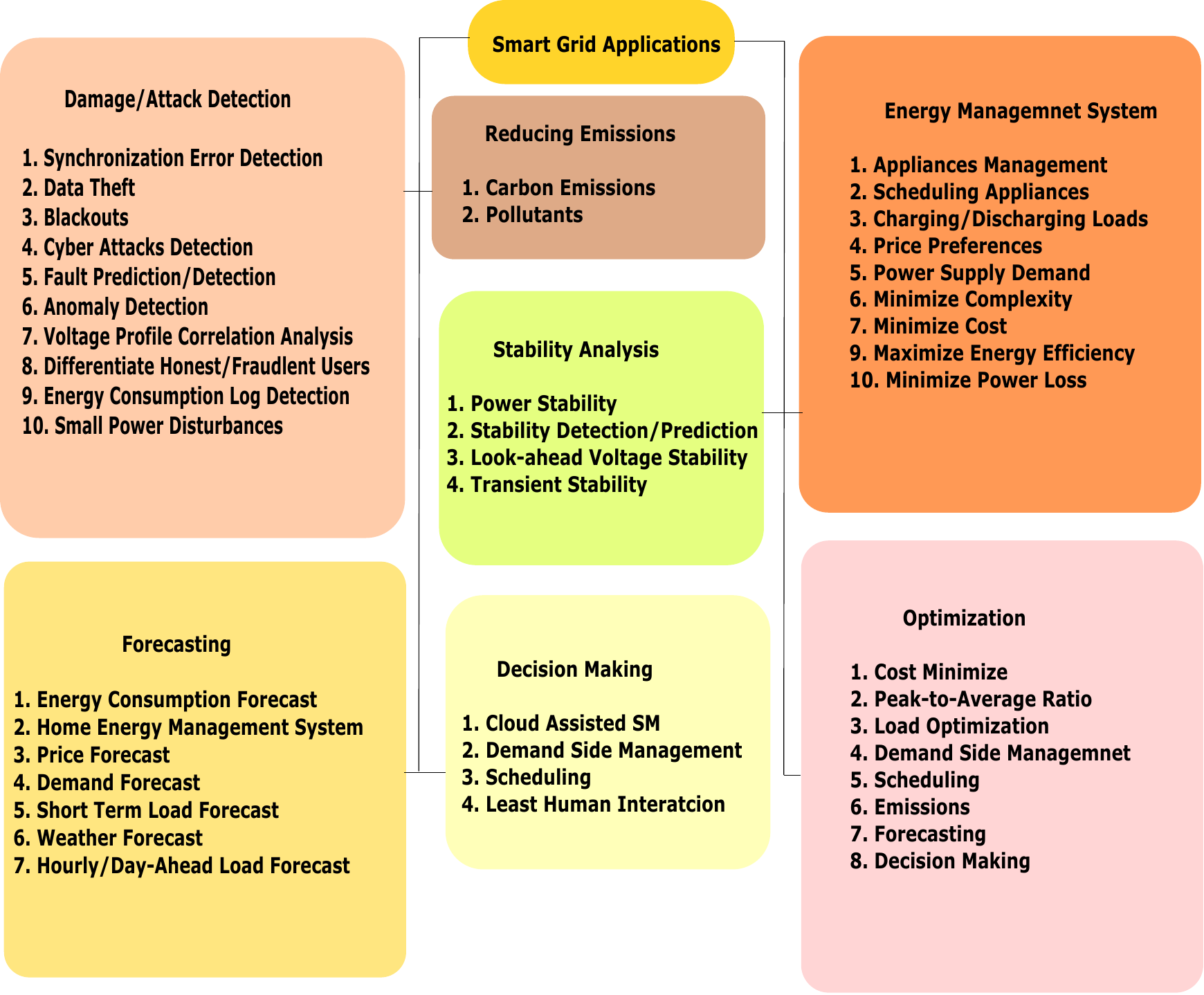}
\caption{Applications of smart grids }
\label{sg_applications}
\end{figure}

The increasing investment in market for smart grid deployment is due to its vast applications and benefits in real-life. The benefits include ease of use, scheduling of devices, economical for utility and users, environmental sustainability, etc. as shown in Figure~\ref{advsg}. These benefits are elevated when Artificial Intelligence (AI), optimization algorithms, forecasting techniques are applied to SGs. AI is now being used in almost every market/industry due to its efficiency, least time consumption in performing tasks, and reduced human involvement. Optimization in SGs helps to achieve sustainable development goals by providing durable and low cost solutions. Forecasting techniques applied in SGs helps in use of more renewable energy and energy management. Some applications of SGs include stability analysis, decision making, damage/attack detection, forecasting, reducing emissions, energy management, and optimization. Figure~\ref{sg_applications} shows the novel applications of SGs discussed in our review. In the following paragraphs, we provide evidence of features and applications of SGs in more details.

\textbf{Stability Analysis} in power grid refers to reliable, continuous, sufficient (according to demand), secure, lower cost (for utility and users), and uninterrupted (no voltage fluctuations) power supply to the consumers~\cite{hassan2013current}. Conventional power grids does not support state-of-the-art stability analysis as compared to SGs. Stability analysis requires power system to be stable in terms of frequency and voltage to maintain equilibrium and provide efficient power supply~\cite{6158623}. W. Hu~\textit{et al.} provided two methods, Aggressive Support Vector Machine (ASVM) and Conservative Support Vector Machine (CSVM), to improve the detection of stability with improved accuracy in power grids~\cite{hu2019real}. Twin Convolutional Support Vector Machine~\cite{8540347} is another way for improving the stability. Another stability prediction hybrid method was developed using Trajectory Fitting (TF) and Extreme Learning Machine (ELM)~\cite{tang2018hybrid}. The ELM-based method is implemented on utility area where as TF is used at local utility (local power station). Correlated Variables were used to note the look-ahead voltage stability in power systems~\cite{nejadfard2022data}. Furthermore, for transient stability in  Phasor measurement units (PMUs), a decision tree based approach was used in~\cite{behdadnia2021new}. Energy balance function such as Hamiltonian was used for detecting stability and unsuitability in power systems~\cite{machado2019hamiltonian}. Cost-sensitive method was used for computing transient stability~\cite{wang2020transient}. Moreover, J.Q. James~\textit{et al.} proposed frequency dependent load power, stable equilibrium, and evaluated voltage stability in the power systems~\cite{james2017intelligent}.

\textbf{Decision Making} is another AI application used in SGs~\cite{skouby2014smart}. Energy efficient AI in-cooperated applications such as Amazon's Alexa~\cite{alexa}, Samsung Smart things~\cite{sam_smart}, Apple Home~\cite{apple}, Ecobee~\cite{ecobee}, Google home~\cite{app_google}, etc., help in monitoring the smart home and energy consumption by variety of appliances in home. With such deep and automatic monitoring of devices/appliances, energy management with least consumer interaction with devices are facilitated. These applications learn from the behaviors of residents daily life routine and schedule the loads as required by efficient decision making. The study in~\cite{reinisch2011thinkhome} and~\cite{kastner2010using} provides AI-optimized smart home by using knowledge (consumer routine patterns or preferences) based information needed to achieve energy efficiency and user comfort. Moreover, cloud analytic-assisted smart meters were developed using advanced AI for Demand-side management for Smart Homes \cite{chen2019design}. Smart meters/grids uses AI for energy consumption and storage (for consumer and utility) optimization. The decisions are made based on the patterns of energy cost from peak/lean hours of day and preferred scheduling tasks (charging Electrical Vehicle (EV), perform other household tasks) for a day.

\textbf{Damage/Attack Detection and Prediction} is one of the most important application of SGs. A minor damage in power-lines, devices, and/or machinery used in SG could result in tremendous problems. For example, synchronization errors, data theft, interrupted power supply, blackouts, whole power system failures, and also adding of false data (cyber attacks). Hence, it is important to detect the damage/attack location and therefore crises/damage prediction algorithms should be proposed to predict and assess any failure/attack that can happen to ensure timely detection of failures. Accurate and timely detection of damage and attacks in SGs helps in improving the privacy of users and utility as well as it prevents the situation of major blackout of power systems. The damaged/faulty location in grid can be detected using feature extraction method using Wavelet-transform in ML~\cite{shafiullah2017wavelet}. A. Belhadi~\textit{et al.} presented a fault detection (anomaly pattern detection in energy systems) to preserve privacy in SGs using PSO~\cite{BELHADI2021102541}. Another anomaly detection scheme was proposed for false data injection attack detection in large scale SGs~\cite{karimipour2019intelligent}. The authors in~\cite{shabad2021anomaly} proposed  an anomaly detection scheme based on ML. British Columbia hydro system's algorithm uses AI to collect the voltage profile correlation analysis and voltage magnitude comparisons to predict the geographical information system errors~\cite{7098424}. S. Zhang~\textit{et al.} proposed a fault prediction method used in power systems (long-short term memory and support vector machine)~\cite{zhang2017data}. Also, L. Lin~\textit{et al.} proposed a fault/damage detection method caused by storms emergency situations with help of sensors to timely detect the location of fault~\cite{lin2014data}. Furthermore, the authors in~\cite{shihavuddin2021image} used deep learning approach for image-based surface damage detection in renewable grids. Deep learning approach was also used for ideal SGs by damage detection in power lines~\cite{tian2021hybrid}. A. Subasi~\textit{et al.} used intrusion detection system and data mining techniques in SGs~\cite{subasi2018intrusion}. The authors in~\cite{kurt2018online} proposed an online cyber attack detection algorithm using the model-free reinforcement learning (RL) in SGs.

Neural networks were used to detect the false data injection attacks~\cite{wang2019accurate} and for mitigating the cyber-physical false data attacks~\cite{ferragut2017real}. Cyber attacks can also be reduced using optimized fuzzy based approach~\cite{ahmad2012fuzzy} and bloom filter based attack detection~\cite{parthasarathy2012bloom}. A cyber-physical sensor called Industrial control systems Resilient Security Technology (IREST) was developed for anomalies detection due to cyber and physical disturbances~\cite{marino2019cyber}. Electricity theft can be detected and reduced using data analysis~\cite{hasan2019electricity} and deep convolutional neural networks (CNN)~\cite{rouzbahani2020ensemble}. PSO-gated was used to reduce electricity thefts by classifying users into honest and fraudulent classes~\cite{ullah2021hybrid}. L. Zhang~\textit{et al.} used PSO-based Support Vector Machine (SVM) to detect anomaly detection in SGs~\cite{zhang2019anomaly}. I. Parvez~\textit{et al.} proposed a supervised ML technique to detect electricity theft in smart meter data using real electricity consumption data from users~\cite{parvez2016securing}. The authors in~\cite{junior2019low} proposed a mechanism that uses low-voltage and low-cost data (energy consumption statistics) exchange between utility and users to detect small power disturbances in Smart Meters (SM) and cyber cascade failure in SGs~\cite{singh2020detection}.

\textbf{Forecasting} helps to reduce peak electricity load and costs in SGs using demand response management, which is a crucial task that depends on price, load, or renewable energy forecasting. Reliable forecasting methods can save more cost of electricity and power in SGs. A variety of algorithms can be applied in SGs for load and price forecasting. Load forecasting time might be varied from few minutes to many years. Many techniques were proposed for load and price forecasting. In this survey we provide the novel and state-of-the-art forecasting techniques published in recent years. The authors in~\cite{izidio2021evolutionary} forecasted energy consumption series and used Genetic Algorithm (GA) for feature selection and hyper-parameter optimization to improve the accuracy of the mechanism. G. Hafeez~\textit{et al.} in ~\cite{hafeez2020innovative} proposed an integrated framework of Artificial Neural Network (ANN) based forecast engine for home based energy management and energy consumption forecasting using Day-Ahead Grey wolf modified Enhanced Differential Evolution algorithm (DA-GmEDE). Authors in~\cite{ayub2019electricity} applied SVM to forecast electric load in SG. A demand forecasting method was presented that employs support vector regression (SVR) for hour-ahead demand forecasting~\cite{fattaheian2014hour}. M. Ali \textit{et al.} in~\cite{ali2019optimum} proposed short term load forecasting algorithm for analysing and mitigating the disturbances causing overloading and variation in demand profile. In~\cite{bessa2015probabilistic}, authors proposed a 6-hour ahead forecasting algorithm  based on the vector auto-regression framework at residential solar photovoltaic (PV) and secondary substation levels. On the other hand, short-term electricity price forecasting and classification based on price threshold can be accomplished using multi-kernel extreme learning machine (MKELM)~\cite{bisoi2020short}. Similarly, big data can also be used for electricity load and price forecasting in SGs to ensure reliability and electricity management in market based on user demands~\cite{mujeeb2018big}. ML algorithm based on cluster analysis proves to be efficient in improved energy consumption forecasting in SGs~\cite{shchetinin2018cluster}. E. Y. Shchetinin in~\cite{muralitharan2018neural} used CNN, Neural Network based Genetic Algorithm (NNGA), and Neural Network based Particle Swarm Optimization (NNPSO) for energy usage and demand forecasting in SGs. Moreover, for probabilistic load forecasting, deep ensemble learning based method was used to classify the customers based on the load uncertainties~\cite{yang2019deep}. By integrating Feature Engineering (FE) and modified fire-fly optimization (mFFO) algorithm with support vector regression (SVR), the authors of~\cite{hafeez2021novel} proposed FE-SVR-mFFO as a forecasting framework in SGs. FE-SVR-mFFO was used for stable and reliable load forecasting for effective planning and secure SG decisions. L. Li \textit{et al.} used deep learning based short-term forecasting (DLSF) method for data clustering using a deep Convolutional Neural Network model for electricity load forecasting~\cite{li2017everything}.

\textbf{Reducing Emissions} is referred to reducing the carbon emissions and pollution. Renewable energy, optimization methods, and energy management processes are widely used to reduce pollutants in the environment. In~\cite{lau2014optimization}, authors used an Ensemble Based Closed-Loop Optimization Scheme (EnOpt) to reduce carbon emissions in SGs. The scheme used Ensemble Kalman Filter (EnKF) for forecasting the carbon emissions during the process of energy generation and consumption. The resultant forecasted values of carbon emissions were optimized using the EnOpt optimization. Authors in~\cite{saber2011resource} used PSO to minimize cost and emissions in renewable and plug-in vehicles. Multi-objective wind-driven optimization (MOWDO) algorithm and Multi-objective Genetic algorithm (MOGA) were used for reducing emissions and operational costs in SGs~\cite{ullah2021multi}. DSM with reduced emissions were obtained using day-ahead load shifting  and Evolutionary Algorithm (EA) for optimized results~\cite{logenthiran2012demand}. Wireless communication technology can be used to manage domestic renewable generation and power shedding algorithm for reduced carbon emissions~\cite{elkhorchani2016novel}. Carbon emissions can be reduced using  complex multi-functional control unit proposed in~\cite{gorbe2012reduction}.

\textbf{Energy Management} in SGs plays significant role for reducing energy loss and cost of energy for both utility and users. SGs are advanced electricity network that can monitor the demand and supply of electricity and manage the tasks to meet the  energy requirements of users at different time periods. Energy Management Systems (EMS) and DSM are two important schemes to overcome the problem of energy management in grids. Following are some popular novel approaches for energy management system in SGs. MATLAB-Simulink-GUIDE tool boxes were used for energy management and house hold appliances management in SGs~\cite{lopez2018smart}. H. Taherian~\textit{et al.} in~\cite{taherian2021optimal} proposed hybrid model for scheduling of home appliances with the help of electricity usage patterns of users of smart meters. A. Manzoor \textit{et al.} proposed hybrid technique Teacher Learning Genetic Optimization (TLGO) for DSM in SGs~\cite{manzoor2017intelligent}. Mixed Integer Linear Programming (MILP) was used to schedule smart appliances, charging/discharging of EVs for energy management, and energy cost reduction~\cite{aslam2020towards}. F. Kennel \textit{et al.} used Hierarchical Model Predictive Control (HiMPC) for SGs~\cite{kennel2012energy}. A consensus-based algorithm was developed for energy management in SGs~\cite{7387777}. A Stackelberg game-theoretic framework was proposed in~\cite{7948731} for energy management  in generators and MGs. Adaptive Fuzzy Logic Model (AFLM)  and Supervised Fuzzy Logic Learning (SFLL) were used to detect the user preferences and pricing for energy management~\cite{keshtkar2017adaptive}. A Hybrid Stochastic/Robust (HSR) optimization method  was  developed by~\cite{9187733} for overall cost minimization in SGs using energy management. Distributed optimization algorithm was used for event driven energy management~\cite{8386427}. To reduce energy expenses in residential area, a scheduling mechanism based on Markov Decision Process (MDP) was presented in~\cite{misra2013residential}. W. Meng and X. Wang proposed Distributed Energy Management Algorithm (DEMA) based on non-convex optimization problem and lagrange duality method in SGs~\cite{meng2017distributed}.

An IoT and cloud-based energy management system was proposed that used the load profile of customers and enables the utility to adapt their energy consumption demand as well as to manage the power supply accordingly~\cite{hashmi2021internet}. P. Duan~\textit{et al.} in~\cite{duan2020distributed} used distributed multi-layer cloud-fog computing architecture energy management  for plug-in hybrid EVs. P. Duan~\textit{et al.} also used GWO and Monte Carlo simulations for increasing efficiency of demand management in SGs. A generic hierarchical architecture using entity abstraction layer and communication abstraction layer was proposed for energy management~\cite{mauser2015organic}. An automation of renewable energy sources with energy management system was proposed using Proportional-Integral (PI), Adaptive Neuro Fuzzy Inference system (ANFIS), and nickel metal hydride battery for storing of excess energy~\cite{subha2021design}. A dynamic energy management for smart-grid powered Coordinated Multi-Point (CoMP) was proposed using  virtual-queue-based relaxation and the stochastic dual-sub gradient techniques for minimizing energy cost with maximum energy management~\cite{7389362}. The authors in~\cite{hurtado2015smart} proposed an advanced Building Energy Management Systems (BEMS) based on multi agent system using PSO to maximize the comfort by minimizing data management complexity and maximizing the energy efficiency.

\textbf{Optimization} in SGs aims at considering cost reduction. Monte Carlo simulation and krill herd optimization algorithm were used for cost minimization in hybrid plug-in vehicles~\cite{rostami2015expected}. An Artificial Immune Algorithm (AIA) was proposed using Pareto optimal solution to solve multi-objective problem. That is, by minimizing operation cost and peak to average ratio at utility end~\cite{8186194}. Integer Linear Programming (ILP) technique is used to minimize the peak hourly load for optimized daily load schedule~\cite{6175785}. A Local-Optimization Emergency Scheduling scheme was developed using local geographic information for large scale-IoT in SGs~\cite{7951030}. L. Xu~\textit{ et al.} in~\cite{xu2018robust} used cyber-physical system robust routing model to evaluate the information flow using the Big-M method. Y. Zhang~\textit{et al.} proposed a trust system using set packing algorithm for optimizing the locations of the trusted nodes, achieving fault tolerance, and finding least cost using optimal routing algorithm in the multiple layer architecture of SG~\cite{zhang2013trust}. The authors in~\cite{7011561} proposed an Evolutionary Algorithm for optimized DSM in SG. L. Liu and Z. Han in~\cite{7069405} used Bayesian network and time of use for load control power management algorithm. Tabu Search optimization algorithm was applied to minimize the generation cost and maximize the project earned by utility using controllable loads (heating pumps, EV, etc.) at demand side~\cite{HOWLADER2016107}. A solution to mix-integer non-linear programming problem was proposed for optimization of centralized robust/stochastic in microgrids plugin-EV for active and reactive powers exchange~\cite{SAFFARI2019105847}. In~\cite{SHAKOURIG2017171}, authors used multi-objective MILP for minimizing electrical peak load and electricity cost for DSM in SG. A parallel distributed optimization approach was used to minimize the complexity, cost, and time of communication using renewable distributed generators~\cite{6576918}. A cloud-based SG information management and network resource optimization models were used to minimize the total cost using MILP and linear programming, respectively~\cite{6471813}. 

Distributed economic dispatch mechanism based on finite time average consensus algorithm and projected gradient was introduced to solve  optimization cost functions from quadratic and non-quadratic convex problems using wind and thermal generators~\cite{7120161}. A distributed optimization algorithm using centralized and de-centralized model predictive control was proposed to minimize the variance in power supplied to residential users~\cite{7399348}. A priority based  optimized traffic scheduling approach was proposed for cognitive radio network based SG for monitoring and modeling of data (command controls, meter readings, and multimedia sensing data) using GA~\cite{6459560}. A model predictive control technique was used for optimal control strategy to optimize the central air-conditioning system and to minimize the expected power according to user demand response~\cite{TANG2019873}. The authors in~\cite{7046415} proposed customer incentive pricing using demand response and GA for resource allocation and maximizing the profit of utility. An optimization model for self scheduling of a hydro-thermal generating utility was proposed for least complexity and more flexibility to be used by decision maker (Genco)~\cite{SOROUDI2013262}. Z. A. Khan~\textit{et al.} proposed meta-heuristic based HEMS using Enhanced Differential Evolution (EDE) and Harmony Search Algorithm (HSA) to schedule the smart appliances and optimization of energy consumption~\cite{khan2019hybrid}. Energy management model was proposed using generated energy, consumed energy information, and MILP to minimize the electricity cost in a smart home~\cite{melhem2018energy}. I. Atzeni\textit{ et al.} focused on optimization and joint optimization problems for demand side distributed energy generation and storage to provide optimal generation and storage suggestions for users using day-ahead optimal scheduling~\cite{6466429}. Glow-worm Swarm Optimization (GSO) and SVM were used to minimize the cost of energy for end-users by load scheduling, rescheduling, and electricity tariff~\cite{puttamadappa2019demand}. 

\subsection{RQ2: Why Smart Grid is a crucial and promising research area?}

\begin{figure}[t]
\centering	
\includegraphics[width=0.7\columnwidth]	{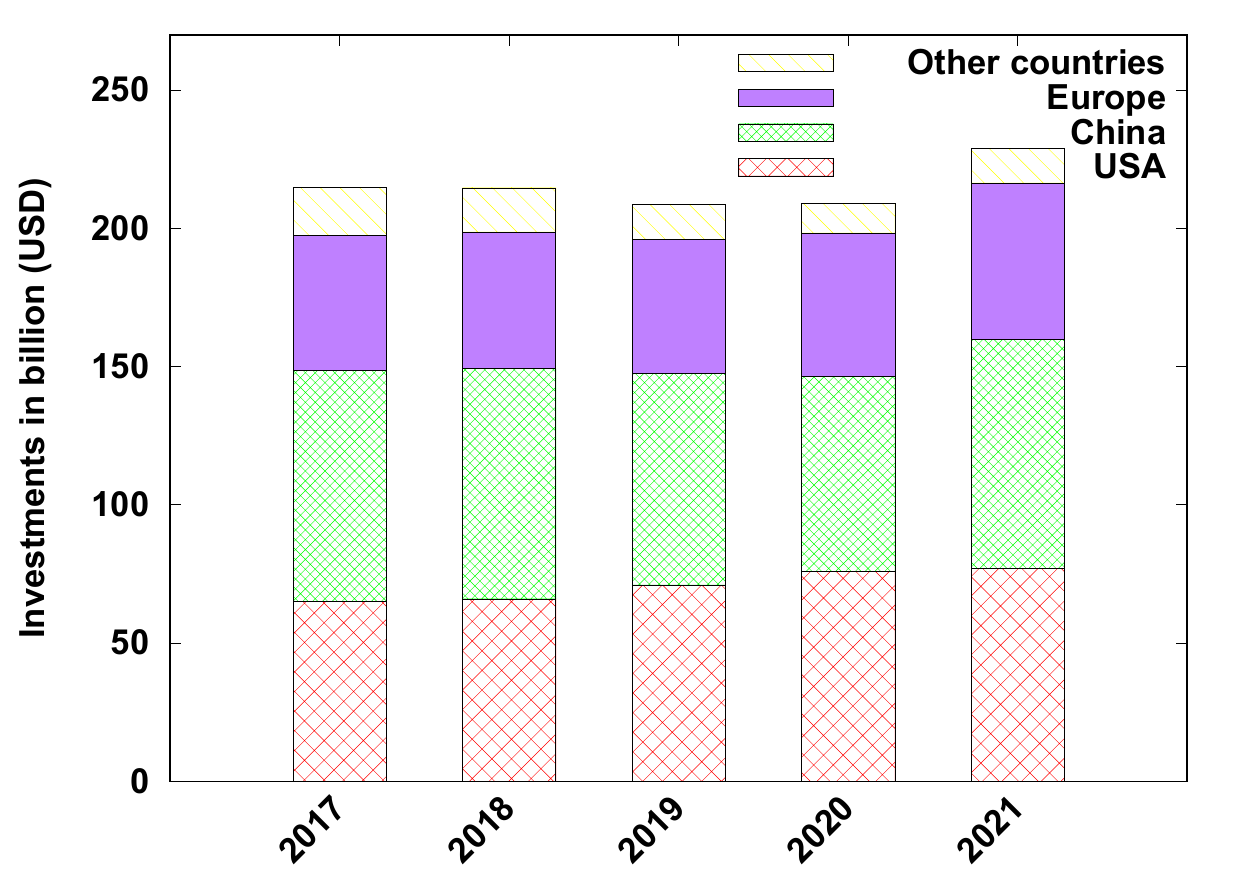}
\caption{Past annual investments in smart grids and smart meters}
\label{SM_SG_STATISTICS}
\end{figure}

SGs provides connectivity to users and utility with a wide range of applications and benefits. SGs technology includes electrical power supply grids, renewable resources, advanced metering, lower emissions, and increased energy efficiency. Due to the benefits in both environmental and financial factors, SGs have become a crucial topic of interest for researchers. Many regions around the world including America, Europe, and Asia have emerged as fastest growing SGs market to manufacture, buy, sell, and invest in modern electrical communication and supply between the utility and users. According to statistical reports~\cite{smartgridstatistics}, since 2017 the annual investments in electricity networks (SGs and smart meters) have increased around the world, as shown in Figure~\ref{SM_SG_STATISTICS}. These statistics verify the higher American, Chinese, and European countries' investments in this sector as compared to other countries around the world. These investments have reached $\sim$56.7 billion USD for Europe, $\sim$82.6 billion USD for China, and $\sim$77.1 billion USD for USA. A total of 115 million smart meters were installed during 2021 in USA (108 million more smart meters units compared to 2007)~\cite{smartgridica}. Furthermore, the smart grid technology market size around the world has increased from $\sim$36.9 billion USD (2021) to $\sim$40.1 billion USD (2022), and it is anticipated to reach $\sim$55.9 billion USD by 2026.

\subsection{RQ3: What are the objective functions/ parameters used for obtaining an optimal energy management/cost-efficient system in Smart Grids?}

\begin{figure}[t]
\centering	
\includegraphics[width=0.7\columnwidth]	{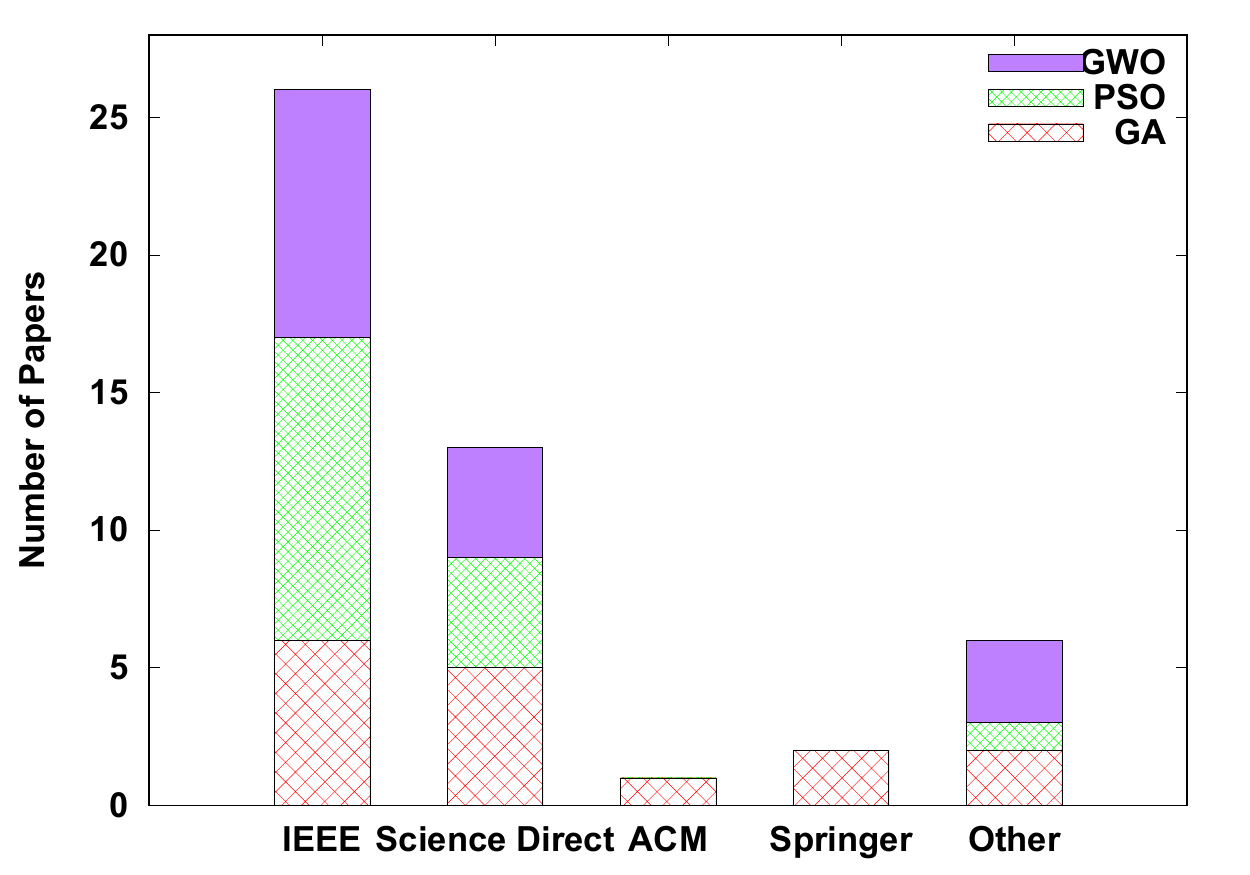}
\caption{Research articles included (per year) from  research resources}
\label{publications_per_year}
\end{figure}
\begin{figure}[t]
\centering	
\includegraphics[width=0.8\columnwidth]	{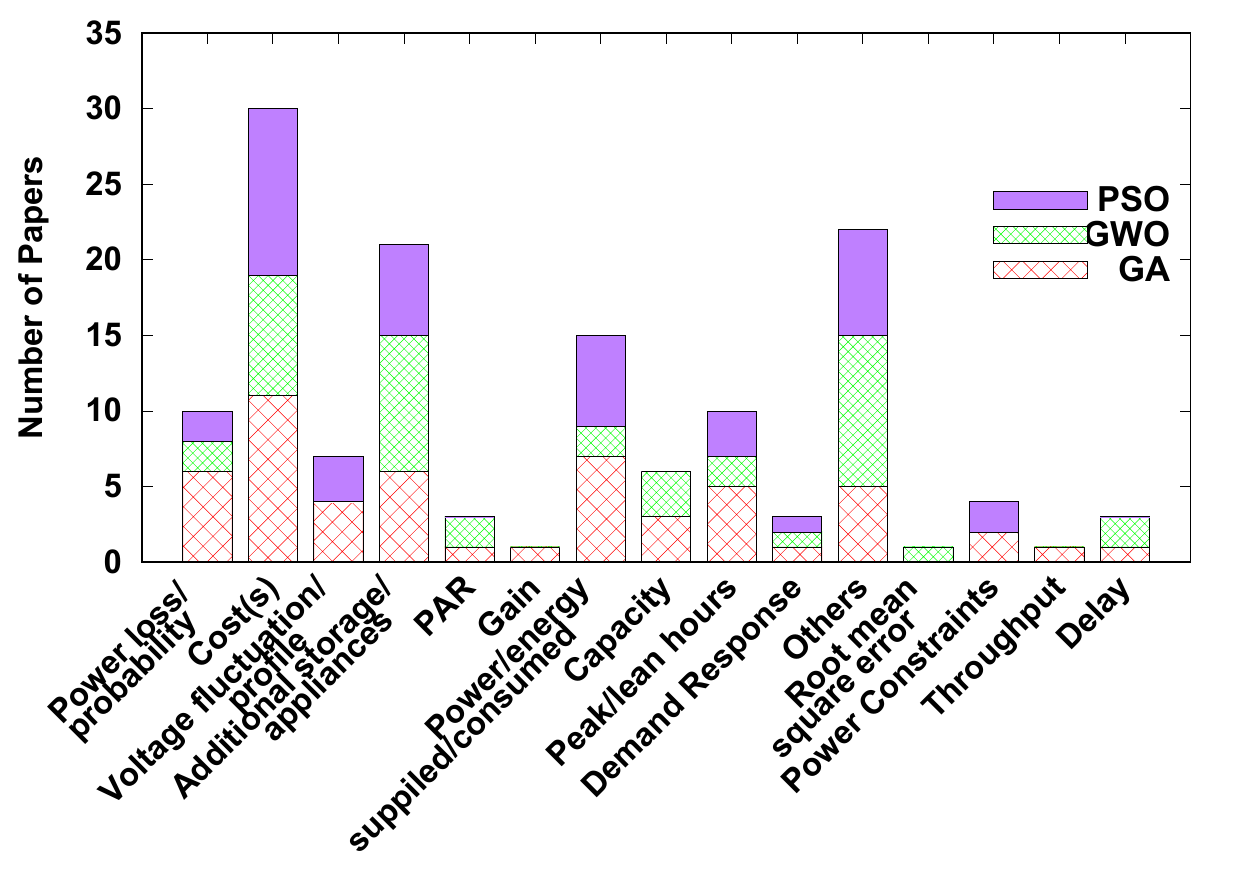}
\caption{Parameters used by the optimization techniques}
\label{parameters_algorithms}
\end{figure}

Tables~\ref{GENETIC},~\ref{GREY}, and~\ref{PSO1} present the objective functions and parameters used for obtaining cost efficient optimal energy management using GA, GWO, and PSO, respectively. Moreover, Table~\ref{PSO1} elaborates the articles that uses PSO to reduce the cost of energy by using energy management techniques in SGs. The collected research articles conclude that energy cost can be minimized by scheduling/managing the loads and optimizing the power losses, energy costs, and utilizing extra energy storage (if used). However, RMSE and peak average ratio are the least used parameters for optimizing the costs with energy management.  We illustrate in Figure~\ref{publications_per_year} the research databases (and the number of articles per each database) for the included articles in addressing RQ3. Figure~\ref{parameters_algorithms} shows details of parameters used by the articles included in this survey. The figure concludes that cost functions are most widely used to optimize the functions. \textit{``Others''} parameter in Figure~\ref{parameters_algorithms} include time stamps, time functions, current, accuracy of load, droop parameters, emissions, transformer and capacitor parameters, latency, etc. As illustrated in Figure~\ref{parameters_algorithms}, the least used parameters to optimize the cost and energy management issues in SGs are Gain and Throughput. 

\subsection{RQ4: How much GA, GWO, and PSO algorithms have been employed to enhance energy and cost handling in Smart Grids?}

\begin{figure}[t]
\centering	
\includegraphics[width=0.7\columnwidth]	{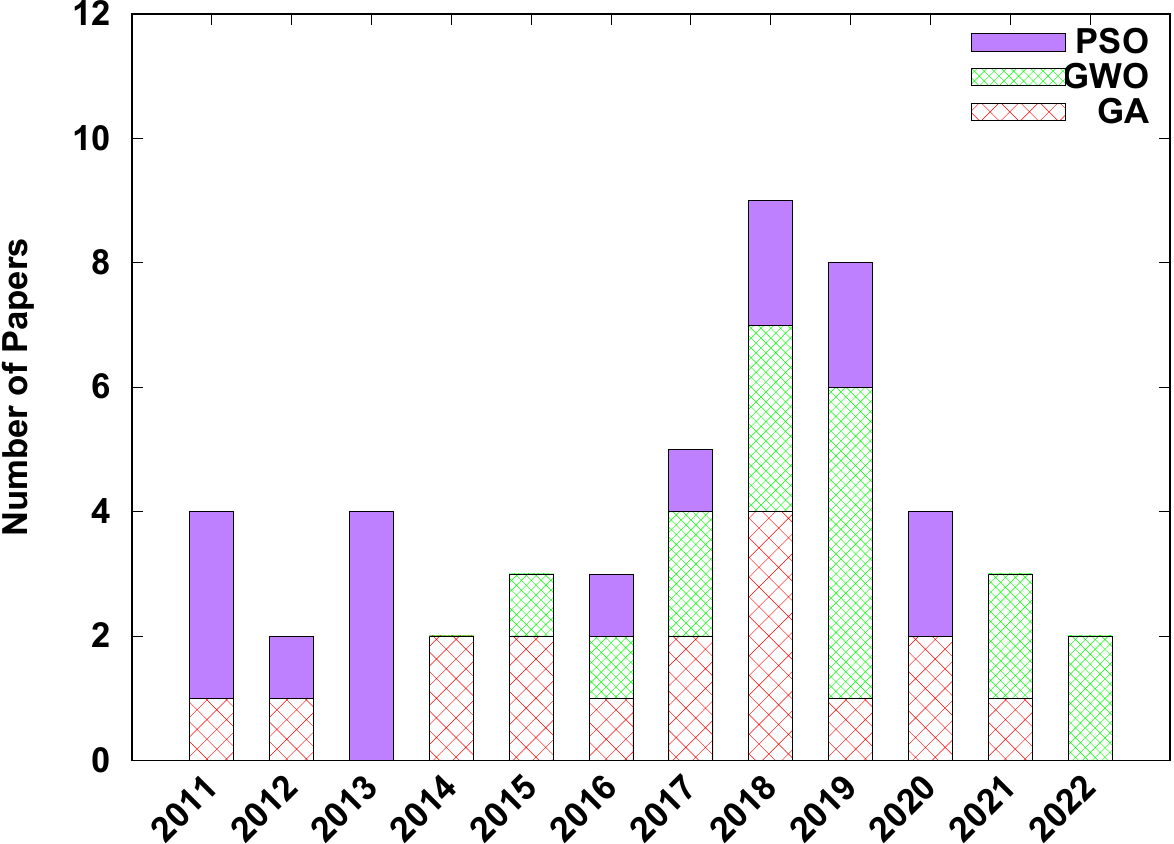}
\caption{Research articles using the algorithms per year}
\label{algopub}
\end{figure}

\begin{figure}[t]
\centering	
\includegraphics[width=0.8\columnwidth]	{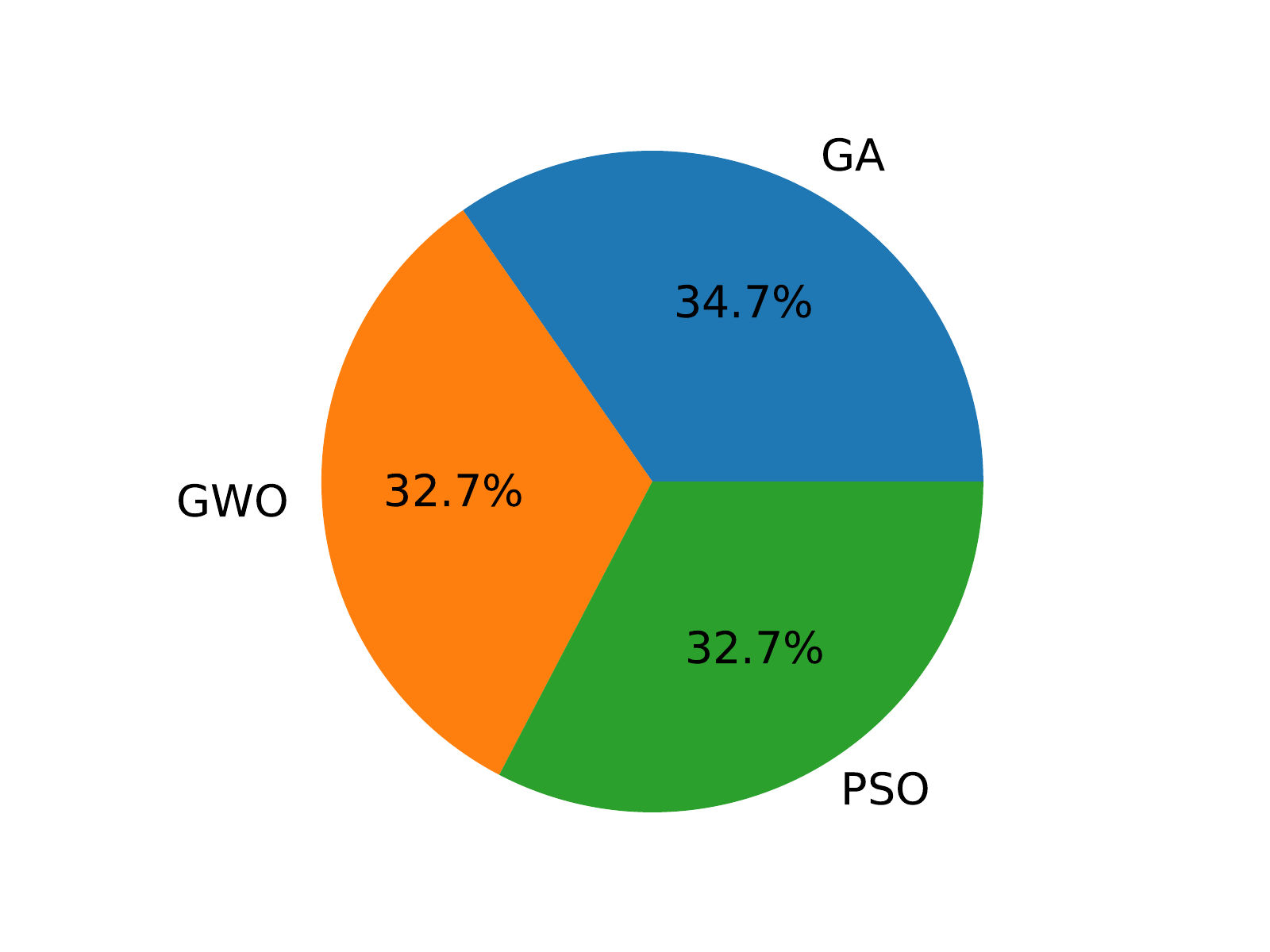}
\caption{Research articles included per optimization algorithm }
\label{algorithms_pie}
\end{figure}

Figure~\ref{algopub} shows the number of publications per year using PSO, GA, and GWO for optimizing the cost and energy in SGs. Furthermore, Figure~\ref{algorithms_pie} shows a pie chart with the percentages of included articles in this survey based on their usage of optimization algorithms. As shown in the aforementioned figures, there is slight more articles used GA as optimization algorithm as compared to PSO and GWO.

\begin{table*}[htbp]
\caption{Optimization using Genetic Algorithm in Smart Grids}
\label{GENETIC} 

\setlength{\arrayrulewidth}{0.5mm}
\setlength{\tabcolsep}{2pt}
\renewcommand{\arraystretch}{1}

\begin{tabular}{|p{1cm} p{8.5cm} p{8.5cm}|} 

    
   \hline
    \textbf{Article}   & \textbf{Objective Function(s)/Motivation} & \textbf{Parameter(s)/Constraint(s)}\\ \hline
\cite{MOZAFAR2017627} &  Location and capacity of renewable energy sources
& Power losses, voltage fluctuations, energy supplying, and  car battery maintenance costs
 \\
   \hline
\cite{PROVATA2015782}   &  Minimizing total energy cost, prediction of power generation and consumption, optimization of modeled micro grid    & Energy cost, and maintenance cost of the energy storage
  \\\hline
\cite{asgher2018smart}  & Minimizing the energy cost and user discomfort & Total cost, per-hour cost, and Peak to Average Ratio (PAR)\\
   \hline
\cite{LEONORI2020105903}       &  Maximizing the profit based on time of use energy price policy
   &   Profit generated by the energy exchange with the grid (time slots and energy costs)\\
   \hline
\cite{MENG2018215}  & Supporting the retailer to make optimal day-ahead dynamic pricing decisions for different customers,  minimizing the payment bill for an acceptable energy consumption,
       & Demand/number of consumers, appliances/usage time, energy consumed, and  cost of energy\\
       \hline
\cite{10.1145/3320326.3320329}  & Optimal charging/discharging
schedule of EVs (Battery Electric Vehicles,Plug-in Hybrid Electric Vehicles) with pricing periods
 &   Daily mileage, charger/power capacity, and low/high power price periods  \\

   \hline
\cite{ABUELNASR20182807}        &    Minimizing the micro-grid energy loss, total cost of imported energy, and carbon-dioxide emissions &  Load generation profiles, energy storage charging/discharging schedule, demand response decisions, dis-patchable and non-dis-patchable
DGs, switching capacitor, and  imported energy from the
main grid\\
\hline


\cite{en7042449}   &  Flattening the load profile, peak load, avoiding power system elements aging by charging and scheduling of EVs &   Costs of power system investment, thermal line
limits, the load on transformers, voltage limits, and charging and parking availability patterns\\
 \hline

\cite{bibi2017running}  &  Minimizing running cost functions in SGs &    Energy cost from utility grid, load Shedding cost, frustration cost, dissatisfaction cost, tariff hour, battery usage, and financial gain of user \\
  \hline

\cite{doi:10.1080/08839514.2016.1138794}   &  Minimizing the network cost of SGs using QoS-Aware routing protocol  &  Packet loss, path loss, effective throughput,
network availability, critical network, packet transmission
delay, and connection outage probability\\
 \hline
\cite{5874894}    &   Minimizing voltage control and active power loss & Real \& reactive power equality constraints, power limits, voltage limits, capacitor step limits, and transformer tap ratio limits \\\hline
\cite{nakabi2018computational}  &   Finding best prices to  maximize the retailer profit  by DSM and demand response programs & economic dispatch, customer behavior learning, price, and load per hour \\
 \hline
 
\cite{bourhnane2020machine}   & Energy consumption scheduling and prediction & Timestamps, power consumption, and  power loss \\ \hline


\cite{7325565}    & DSM and to regulate voltage, and Synchronization of smart load and power generation profile & voltages, current (AC/DC), time-domains, and power\\ \hline


\cite{wang2021example}  &  Minimizing total operating cost, and maximizing profit of SGs
 &  Power, time, and cost \\ \hline

\cite{yan2014impacts} & Energy management by shifting the load of EVs to reduce the cost & Time, power consumption, power loss, and price\\\hline

\cite{roche2012framework} &  Deployment of feeder buses for distributed generation at least cost and the highest reliability & Un-served energy, and costs
   


\\
\hline

\end{tabular}
\end{table*}

\begin{table*}[htbp]
\caption{Optimization using GWO in Smart Grids}
\label{GREY} 
\setlength{\arrayrulewidth}{0.5mm}
\setlength{\tabcolsep}{2pt}
\renewcommand{\arraystretch}{1}
\begin{tabular}{|p{1cm}p{8.5cm}p{8.5cm}|}

   \hline
    \textbf{Article}   & \textbf{Objective Function(s)/Motivation} & \textbf{Parameter(s)/Constraint(s)}\\ \hline
\cite{kalkhambkar2016joint} & Minimizing energy loss by optimal allocation of energy storage & Battery constraints, voltages, feeder current capacity, and power balance\\
   \hline
   
 \cite{kumar2021novel}  &  Energy management &  Cost functions, power resources,  charging, and discharging patterns 
   \\\hline
   
   
\cite{yin2022distributed}  & Optimization to reduce unit cost and carbon emissions &  Active power, carbon-dioxide emissions with generation cost \\
   \hline
   
\cite{huang2022building}  & Energy management &   Power generation, consumption, battery charging, and discharging \\\hline
       
\cite{mahdad2015blackout} & Prevention of blackouts and minimizing costs & Power generation,
 voltage, shunt capacitors/reactors, and transformers tap ratio  \\
  \hline
  
\cite{makhadmeh2021novel}     &  Power scheduling  &   Capacity power limit rate, waiting time, power, and peak hours \\\hline

 \cite{makhadmeh2018optimal} & Power scheduling using smart battery &   Cost, peak to average ratio, power, and  waiting time of appliances\\
 \hline

\cite{moazami2018optimal}  & Minimizing power losses, regulating voltage stability & Distributed generation droop parameters, location and capacity of distributed generation units, capacitors, transmission power-lines, and renewable energy resources
\\
 \hline
 
\cite{mostafa2017performance}  &   Minimizing generating costs, emission reduction & Fuel costs, power loss, and emission\\
  \hline
  
 \cite{molla2019integrated}   &  Energy management system and cost reduction  & Cost, power generation, and time of energy usage \\
 
\hline
  
 \cite{gazijahani2017optimal} &  Power scheduling of MGs, minimizing operational and power loss costs, reducing pollutants emissions,  Power loss cost & load/demand energy, cost, and energy storage \\
 \hline
  
  

   \cite{azizuddin2019optimum} &  Optimizing parallel
DC (direct current)-DC converters to distribute the load demand between different converters  & RMSE, delay, voltage,current, and time \\
\hline


   \cite{naz2019game}  &  Reducing the operational time using energy management system & Peak average ratio, energy consumed, generated and load pricing \\
\hline


 \cite{mostafa2018application} &   Minimizing
the generation costs and emission reduction & Gas emissions, power loss, fuel, and total cost   \\
\hline

 \cite{cao2019energy}    &   Energy management
of an isolated micro-grid and distributed energy resources &  Gas emission, charging and discharging batteries, maintenance, and operational cost  \\
\hline

 \cite{naz2019short}   &   predicting the price  and load  & Accuracy of load, and price of electricity  \\\hline  

\end{tabular}
\end{table*}

\begin{table*}[htbp]

\caption{Optimization using PSO in Smart Grids}
\label{PSO1} 
\setlength{\arrayrulewidth}{0.5mm}
\setlength{\tabcolsep}{2pt}
\renewcommand{\arraystretch}{1}

\begin{tabular}{|p{1cm}p{8.5cm}p{8.5cm}|} 

 \hline
    \textbf{Article}  & \textbf{Objective Function(s)/Motivation} & \textbf{Parameter(s)/Constraint(s)}\\ \hline
   
 \cite{faria2011demand}    &  Minimizing the costs at consumers-end using demand response management     &  Total consumption cost, pricing hours and cost, and consumed energy
   \\
   \hline
   
\cite{soares2013day}    & Energy resource management and day-ahead scheduling of grids using electric vehicles   &  Voltages, thermal limits, battery storage limits, charging and discharging of batteries, and  energy generation cost \\
   \hline
   
   
       
  
\cite{soares2016multi}      &  Energy resource management and minimizing electricity acquisition costs &   Active and reactive power generation, charging and discharging of battery capacity, time, and costs\\
\hline

 \cite{steer2012decision}      & Minimize the total
energy input  &   Total required input
energy  and energy lost in transmission, demand deficit, and change in temperature of the transmission medium\\
 \hline

\cite{soares2013application}    & Energy resources management and energy resources scheduling 
  & Costs, energy costs, power, and battery charging/discharging limits
\\
 \hline
 
  
  
 \cite{sousa2018flexibility}  & Equilibrium of users and utility profits and  HEM system & Starting and usage duration time of appliances, and power (baseline and consumed) profile the appliances  \\
 \hline
  
  
   \cite{kasi2020operation}   &  Minimizing the operational cost of
micro grid using combination of renewable resources  & Total energy and operation costs (fuel cost, maintenance and
operation costs of distributed generation ), main-grid operating cost, start-up or shut-down cost of fuel cells, and micro-turbine \\
\hline

   \cite{keles2017multi}  &  Minimizing the daily cost of renewable grids by optimal energy mixing rates, energy balance, and anti-islanding constraints & Output powers, energy mixing rate factor, and electricity cost\\
\hline



  \cite{mechta2020energy}   &   Optimal energy path between utility and users to minimize the energy and cost loss & Path latency, cost, and path quality   \\
\hline

 \cite{faia2019demand}   &  Demand response optimization methodology for generic residential house to minimize the daily costs  & Buying and selling cost of energy, power consumption, and contracted power costs  \\
\hline

 \cite{vale2011lmp}   &   Energy resource scheduling & Energy market price, load, power generated, daily time and date, and weather forecasting  \\
\hline


 \cite{faria2013modified}  &  Minimizing the operation costs   &  Power generation, bus voltages, active and reactive power generation units, and thermal limits   \\
\hline

   \cite{farhadi2019smart}  &   Maximizing the profit & Maximum/minimum output power and power balance, ramp up/down rate, battery charging/discharging, and  loads information  \\
\hline


  \cite{faria2011particle}   &  Minimizing the operation costs by demand response scheduling
& Quadratic energy cost functions, power, and load   \\
\hline




   
   
\cite{herath2018computational}    & Price responsive
demand flexibility for advanced metering infrastructure (micro-grid)  &  Energy price (price fluctuations), energy demand, and  energy forecasted values \\
   \hline

 \cite{saif2013optimal}   & Optimal allocation and capacity of distributed energy resources to minimize cost  &  Charging/discharging battery, load,  and  energy available/used values \\
   \hline

\hline
\end{tabular}

\end{table*}

\section{Open Research Challenges and Future Directions}\label{opc}

In this section, the analysis of this SLR is presented based on the findings from the results of addressing the research questions. As we discussed earlier, SGs have vast range of features and applications (i.e., scheduling, management, AI, and Internet of Things (IoT)) that have been addressed by researchers. However, as SGs became more applications friendly, they have been influenced by high risks of cyber and physical attacks. These attacks might target SGs' monitoring and modeling as well as the privacy of users (their working hours, timings if users are in home, or the devices used by users to monitor voltage patterns). Scheduling of appliances and managing the energy also raises the concerns of information disclosure, user discomfort, and/or the usage of appliances in peak hours. Forecasting and damage detection are other important common features and application of SGs. Attackers might misuse the forecasting information for future predictions of consumer availability in home, at work, or even power loss from the appliances. In the following paragraphs, we discuss the research challenges modeled after we deeply analyzed the research articles involved in this SLR. These research challenges are categorized into:

\begin{itemize}
    \item Cyber and physical attacks and damage detection.
    \item Monitoring and modeling.
   \item  Which optimization algorithm?
\end{itemize}

\textbf{Cyber and physical attacks} are accomplished when the exchanged data among the utility and users is not/enough secure. Furthermore, in case of an un-trusted third party has the ability to access, read, replicate, or modify the data. Many research works have been conducted to mitigate the cyber attacks in SGs, such as~\cite{shabad2021anomaly,kurt2018online,ferragut2017real,ahmad2012fuzzy,parthasarathy2012bloom}. The proposed intrusion detection systems were designed with the help of data mining techniques, model reinforcement learning, deep learning, and fuzzy-based approaches. Physical attack detection in SGs has the same importance as detection of cyber attacks. Using algorithms that can detect both cyber and physical attacks with one powerful technique would reduce the latency of computing, detection, and processing time to counter/address the attack(s) with reduced costs. However, few research articles like~\cite{wang2019accurate} employed neural networks cyber-physical false data attack detection. D. L. Marino \textit{et al.} in~\cite{marino2019cyber} proposed cyber-physical attack detection sensors to reduce the anomalies. Therefore, the ratio of research works conducted for detection, prevention, or mitigation of both cyber and physical attack at same time is less when compared to cyber or physical attack detection.

Wide area network (WAN) is used for data exchange and is physically accessible to everyone. The attackers who pretending as legitimate users might also access the data flow through the communication channels in WAN. Hence, the attacker might become burden by introducing DoS attacks or by injecting fake information to forge the billing receipts of legal electricity buyers. Similarly, in wireless mediums, the transmitter nodes, communication channels, and receiver nodes are at high risk of being compromised. In such cases, proposing a secure hardware and implementing effective/secure designs for sensing the attacks must be taken into account by researcher (as compared to the theoretical-based researches). Optimization algorithms should also be used for intrusion detection system to reduce the trade-off between detection cost and providing a secure information exchange between users and utility. By analyzing the techniques for intrusion detection system in this SLR, we conclude that most of the research works in this field are related to theoretical applications and attack detection system. Hence, future research works should consider real data and scenario when proposing the attack/damage detection scheme.

\textbf{Monitoring and modeling} in SGs has vast applications when interconnected to each other, i.e., IoT of SGs. To monitor such a large number of devices or appliances interconnected to the grid at the same time, is quite a challenging task that has a high risk of failure. Conventional monitoring requires data synchronization in SGs, however, with growing demands, more applications have been introduced that need continuous (wide area/long range) connectivity and monitoring. Monitoring is not only limited to data synchronization, but also related to the performance analysis in terms of system service, stability analysis, and securing the energy/data flow. Proper monitoring will lead to more advancements in the field of modeling SGs. AI, big data, data mining, and ML techniques should be widely used for monitoring and modeling of appliances, path, energy, and for electrical-system services. Accurate and timely detection of threats will reduce the gaps between the secure communications and energy lay-backs, and/or below average services. By using all information gathered through monitoring the devices, environment, resources as well as demand, the modeling of a new efficient and effective system can be accomplished by employing Artificial Intelligence applications and/or ML techniques.

\textbf{Which optimization algorithm?} Optimization techniques in SGs have been widely used. As described in RQ4, PSO is the most commonly used optimization algorithm for gaining the desired results. When comparing the two algorithms (GA, PSO), we conclude that PSO has many benefits over GA such as less number of used parameters~\cite{hassan2005comparison}. Hence, PSO is preferred over GA for optimizing the results. GWO  was introduced in 2014; Being a recently developed optimization algorithm, its ease of use, flexibility, and providing most accurate results compared to GA and PSO, GWO is gaining popularity and is widely used. Indeed, GWO is considered as the most effective algorithm due its fast convergence and larger search space.

\section{Conclusion}
This systematic literature review helps in identification of novel applications and quality research works in energy management and importance of adapting SGs in real-world scenarios. We reviewed research works from the past 10 years by following the systematic literature review guidelines proposed by B. Kitchenham and S Charters in~\cite{Kitchenham07guidelinesfor} to improve the results of the review topic and gain high-quality materials. The paper discussed the main applications of SGs (such as stability analysis, forecasting, decision making, damage/attack detection, energy management, reducing emissions, and optimization). Afterwards, this survey addressed different raised research questions and proposed different research topics to be considered and studied by future researchers.  

\bibliographystyle{unsrt}
\bibliography{optimization}

\end{document}